\documentclass{article}
\usepackage{PRIMEarxiv}

\usepackage[
backend=biber,
style=ieee,
citestyle=numeric-comp,
sorting=none,
url=false,
isbn=false,
uniquename=false,
alldates=year,
clearlang=true,
maxbibnames=12,
minbibnames=6
]{biblatex}

\DeclareFieldFormat[article,unpublished,misc,preprint]{title}{``#1''}

\let\oldcite\cite
\renewcommand{\cite}[1]{{\mbox{\oldcite{#1}}}}

\AtEveryBibitem{\clearlist{publisher}}
\AtEveryBibitem{\clearfield{note}}
\AtEveryBibitem{\clearlist{language}}
\AtEveryBibitem{\clearname{editor}}
\AtEveryBibitem{\clearfield{series}}
\AtEveryBibitem{\clearlist{location}}

\usepackage{doi}

\usepackage{booktabs}
\usepackage{siunitx}
\usepackage{fancyhdr}
\usepackage{hyperref}
\usepackage[font=small,labelfont=it]{caption}
\usepackage{graphicx}
\usepackage[dvipsnames, usenames]{xcolor}
\usepackage{soul}

\let\oldsubsection\subsection
\newcommand{\subsecno}[1]{\oldsubsection*{#1}\addcontentsline{toc}{subsection}{#1}}

\usepackage{placeins}
\usepackage{amsmath,amsfonts,amssymb, bm}

\usepackage{lineno}
\usepackage{mathtools}
\DeclarePairedDelimiter\abs{\lVert}{\rVert}

\newcommand{\LCC}{*_b}
\newcommand{\had}{\circ}

\renewcommand{\vec}[1]{\mathbf{\bm{#1}}}
\newcommand{\mat}[1]{\mathbf{#1}}
\newcommand{\T}{^{\intercal}}
\newcommand{\de}{\mathrm{d}}
\newcommand{\appropto}{\mathrel{\vcenter{
  \offinterlineskip\halign{\hfil$##$\cr
    \propto\cr\noalign{\kern2pt}\sim\cr\noalign{\kern-2pt}}}}}

\usepackage{lipsum}
\begin{document}

\title{Distributed Representations Enable Robust Multi-Timescale Symbolic Computation in Neuromorphic Hardware}

\author{\textbf{Madison Cotteret}\textsuperscript{1,2,3,*} \
\textbf{Hugh Greatorex}\textsuperscript{1,2}, \
\textbf{Alpha Renner}\textsuperscript{4}, \
\textbf{Junren Chen}\textsuperscript{5}, \\
\textbf{Emre Neftci}\textsuperscript{4}, \
\textbf{Huaqiang Wu}\textsuperscript{6}, \
\textbf{Giacomo Indiveri}\textsuperscript{5}, \
\textbf{Martin Ziegler}\textsuperscript{7}, \
\textbf{Elisabetta Chicca}\textsuperscript{1,2} 
}

\maketitle

\let\thefootnote\relax\footnotetext{\hspace{-1.8em}\textsuperscript{1} Bio-Inspired Circuits and Systems (BICS) Lab, Zernike Institute for Advanced Materials, University of Groningen, Netherlands.\\
\textsuperscript{2} Groningen Cognitive Systems and Materials Center (CogniGron), University of Groningen, Netherlands.\\
\textsuperscript{3} Micro- and Nanoelectronic Systems (MNES), Technische Universit\"at Ilmenau, Germany.\\
\textsuperscript{4} Forschungszentrum Jülich, Germany.\\
\textsuperscript{5} Institute
of Neuroinformatics, University of Zürich and ETH Zürich, Switzerland.\\
\textsuperscript{6} School of Integrated Circuits, Tsinghua University, Beijing, China.\\
\textsuperscript{7} Energy Materials and Devices, Department of Materials Science, Kiel University, Germany.\\
\textsuperscript{*} Corresponding author: \texttt{m.cotteret@rug.nl}}

% I added this to get page numbers with the current document class (remove when class is changed)
\thispagestyle{plain}
\pagestyle{plain}

%%%%%%%%%%%%%%%% Text of abstract goes here %%%%%%%%%%%%%%%%%%%%%%%%%%

\begin{refsection}

\begin{abstract}
Programming recurrent spiking neural networks (RSNNs) to robustly perform multi-timescale computation remains a difficult challenge. To address this, we describe a single-shot weight learning scheme to embed robust multi-timescale dynamics into attractor-based RSNNs, by exploiting the properties of high-dimensional distributed representations. We embed finite state machines into the RSNN dynamics by superimposing a symmetric autoassociative weight matrix and asymmetric transition terms, which are each formed by the vector binding of an input and heteroassociative outer-products between states. Our approach is validated through simulations with highly nonideal weights; an experimental closed-loop memristive hardware setup; and on Loihi 2, where it scales seamlessly to large state machines. This work introduces a scalable approach to embed robust symbolic computation through recurrent dynamics into neuromorphic hardware, without requiring parameter fine-tuning or significant platform-specific optimisation. Moreover, it demonstrates that distributed symbolic representations serve as a highly capable representation-invariant language for cognitive algorithms in neuromorphic hardware.

\vspace{1em}

\end{abstract}

\section{Introduction}
\label{sec:introduction}

Neuromorphic computing promises to match the efficiency of information processing in the brain by emulating biological neural and synaptic dynamics in hardware \cite{mead_neuromorphic_1990}. In contrast to more conventional artificial neural networks (ANNs), biological neurons have rich internal temporal dynamics and interact sparsely with unary events (spikes). Performing multi-timescale tasks -- such as motor planning and execution -- with biologically-plausible spiking neurons requires information to be retained and processed for periods much longer than the timescales available in individual neurons and synapses \cite{Ganguli_etal08,khona_attractor_2022}. Information must then be represented by the collective dynamics of recurrently-connected populations of neurons instead. Programming recurrent spiking neural networks (RSNNs) to perform long-timescale tasks with short-timescale neurons remains a formidable challenge however, due to the conflicting demands that the network should react quickly to input but otherwise be stable on long timescales \cite{hochreiter_long_1997}.

In the face of these challenges, a possible approach would be to apply gradient-based iterative learning algorithms to train the network parameters to solve a particular task~\cite{bellec_solution_2020, cramer_surrogate_2022, neftci_surrogate_2019}.
However, besides being computationally intensive and biologically implausible, such approaches are not well-suited to tasks with long timescales, due to the need to backpropagate gradients into the distant (potentially infinite) past \cite{miller_stable_2019, bengio_learning_1994}.
Furthermore, these approaches do not have guarantees of stability and robustness, compared to those afforded by much simpler biologically-inspired models like the Hopfield attractor network~\cite{hopfield_neural_1982, amit_modeling_1989}.

If the trained networks are to be deployed on mixed-signal neuromorphic hardware, one must usually perform calibrating chip-in-the-loop training, lest device-specific nonidealities absent during training lead to a catastrophic degradation in network performance upon deployment~\cite{bohnstingl_biologically-inspired_2022, schmitt_neuromorphic_2017, demirag_online_2021, cakal_gradient-descent_2024, cramer_surrogate_2022}.
To overcome these difficulties and to shift to more robust training procedures, it has been suggested that a more appropriate level of description for programming neuromorphic hardware may be at the level of distributed high-dimensional patterns of neuron activity, rather than individual neuron and synapse parameters~\cite{kleyko_vector_2022}.
With vector symbolic architectures (VSAs) -- also known as hyperdimensional computing (HDC) -- arbitrary symbolic data structures can be represented by high-dimensional random vectors, known as \textit{hypervectors}, such that information is distributed across the entire vector~\cite{kleyko_survey_2022, kanerva_hyperdimensional_2009, plate_holographic_1995, gayler_multiplicative_1998}.
These representations are then not dependent upon the precise functioning of individual components, which is a critical property for robust functioning in neuromorphic hardware and biology~\cite{karunaratne_-memory_2020, cotteret_vector_2024,renner_neuromorphic_2024, faisal_noise_2008}.

In this work, we show how distributed representations can be leveraged to program arbitrary state machines into RSNNs in a scalable and robust manner.
For each state, we add an autoassociative outer product to the recurrent weight matrix to store it as a fixed-point attractor in the RSNN~\cite{hopfield_neural_1982}. Additional heteroassociative outer product terms are then superimposed, which are each responsible for storing an input-triggered transition between attractor states.
Neural state machines have already been shown to be important computational primitives for reproducing cognitive behaviours in neuromorphic agents~\cite{neftci_synthesizing_2013}, and have been used in the context of solving constraint satisfaction problems~\cite{liang_neuromorphic_2019}.

We first embed multiple state machines into simulated RSNNs with nonideal synaptic weights.
To validate these simulations in neuromorphic hardware, we create a proof-of-concept implementation using a memristive crossbar setup run in a closed-loop with simulated neurons. Finally, we implement these networks on Intel's asynchronous digital neuromorphic research chip Loihi 2~\cite{orchard_efficient_2021}, to demonstrate the algorithm's scalability.

This work demonstrates the capability of distributed symbolic representations as a functional abstraction layer for neuromorphic hardware~\cite{kleyko_vector_2022}. It permits the description of complex RSNN behaviours from a high level while being invariant to the choice of underlying neural representation, and is highly robust to hardware-specific nonidealities. It is a step towards a general framework for interoperability of cognitive algorithms across varying neuromorphic hardware architectures, without requiring significant optimisation or fine-tuning upon deployment to each particular platform.

\section{Results}

State machines are perhaps the most comprehensive model of multi-timescale dynamics, as they react quickly to input but are otherwise stable on indeterminately-long timescales. They also have a clear utility for building more complex models of cognitive behaviours~\cite{neftci_synthesizing_2013, dayan_simple_2008, liang_neuromorphic_2019}.
We focus on the task of embedding deterministic finite automata (DFAs) into RSNNs by expressing the DFA in terms of high-dimensional attractor states and transitions between them.
For every state $q$ in the finite set of states $Q$, we embed an associated fixed-point attractor state in the RSNN dynamics that is stable on long timescales. When input is given to the network, the RSNN should quickly transition between attractor states according to the desired state transition function $F: Q \times S \rightarrow Q$ in the DFA, where $S$ is the set of all input symbols.

We generate independent hypervectors $\vec{q} \in \{0,1 \}^N$ to represent each DFA state $q$, and store them as attractors in the RSNN using a Hopfield-like outer-product learning rule \cite{amari_characteristics_1989, tsodyks_enhanced_1988, amit_modeling_1989}. The hypervectors follow a sparse block structure, i.e., the $N$ elements are split up into $M$ blocks of equal length $L$, and exactly one entry in each block is nonzero~\cite{laiho_high-dimensional_2015, frady_variable_2023}.

We use sparse rather than dense vectors to benefit from the increased capacity in attractor networks~\cite{knoblauch_memory_2010, amari_characteristics_1989, tsodyks_enhanced_1988} and distributed rather than localist representations for improved robustness properties~\cite{rumelhart_parallel_1986}.
We use a sparse block structure due to the simplicity of the required neural activation function, and compatibility with the VSA framework~\cite{frady_variable_2023,renner2022sparse}. Additionally, we move closer to a more plausible model of biological neural dynamics, where sparse representations enforced by local competitive mechanisms are abundant~\cite{palm_neural_2013, rozell_sparse_2008}.

For every state transition, we embed additional outer product terms such that if the correct subset of neurons is masked by external inhibition, the RSNN will transition between attractor states (see Methods).
The terms are of the form \mbox{$(\vec{q} - \vec{q}_0) (\vec{q}_0 \had \overline{\vec{s}})\T$} where $\overline{\vec{s}}$ is another (bipolar) hypervector, and ``$\had$'' is a Hadamard product, acting as a hypervector binding operation. Without input to the network, these terms are pseudo-orthogonal to the network state, i.e. $\langle (\vec{q}_0 \had \overline{\vec{s}}) \cdot \vec{q}_0 \rangle = 0$ and so have negligible effect. When the network is masked by the correct (binary) hypervector $\vec{s}$ however (effectively an unbinding operation), they become similar to the network state, i.e. $ (\vec{q}_0 \had \overline{\vec{s}}) \cdot (\vec{q}_0 \land \vec{s} ) = \frac{1}{2}M$ where ``$\land$'' is component-wise $\mathrm{AND}$, implementing the masking. This allows us to effectively modulate the strength of superimposed terms by masking neurons in the RSNN~\cite{cotteret_vector_2024}.

We then wish to program this weight matrix onto neuromorphic hardware to reliably and robustly realise the desired dynamics without requiring significant optimisation or adaptation to the target hardware. In contrast, deployment of RSNNs to neuromorphic hardware usually requires some degree of chip-in-the-loop training or post-deployment fine-tuning to ensure robustness to hardware-specific nonidealities~\cite{schmitt_neuromorphic_2017, bohnstingl_biologically-inspired_2022, cramer_surrogate_2022, neftci_device_2010, buchel_supervised_2021}. This requirement severely limits the ability of neuromorphic hardware to be deployed at scale.

We demonstrate the reliability and generality of our approach by running experiments in three SNN environments. First in simulation, where we show the ideal functioning of the RSNN even with considerably nonideal weights. Second, a closed-loop memristive hardware setup with 64$\times$64 resistive RAM (RRAM) devices acting as the synaptic weights, demonstrating the suitability of the approach for novel beyond-CMOS memristor-integrated neuromorphic hardware. Third, we demonstrate on Intel's neuromorphic research chip Loihi 2 that the approach scales up seamlessly to large state machines.

\subsection{SNN simulation demonstrates robustness to noisy weights}

\begin{figure*}
\centering
\includegraphics[width=0.9\linewidth]{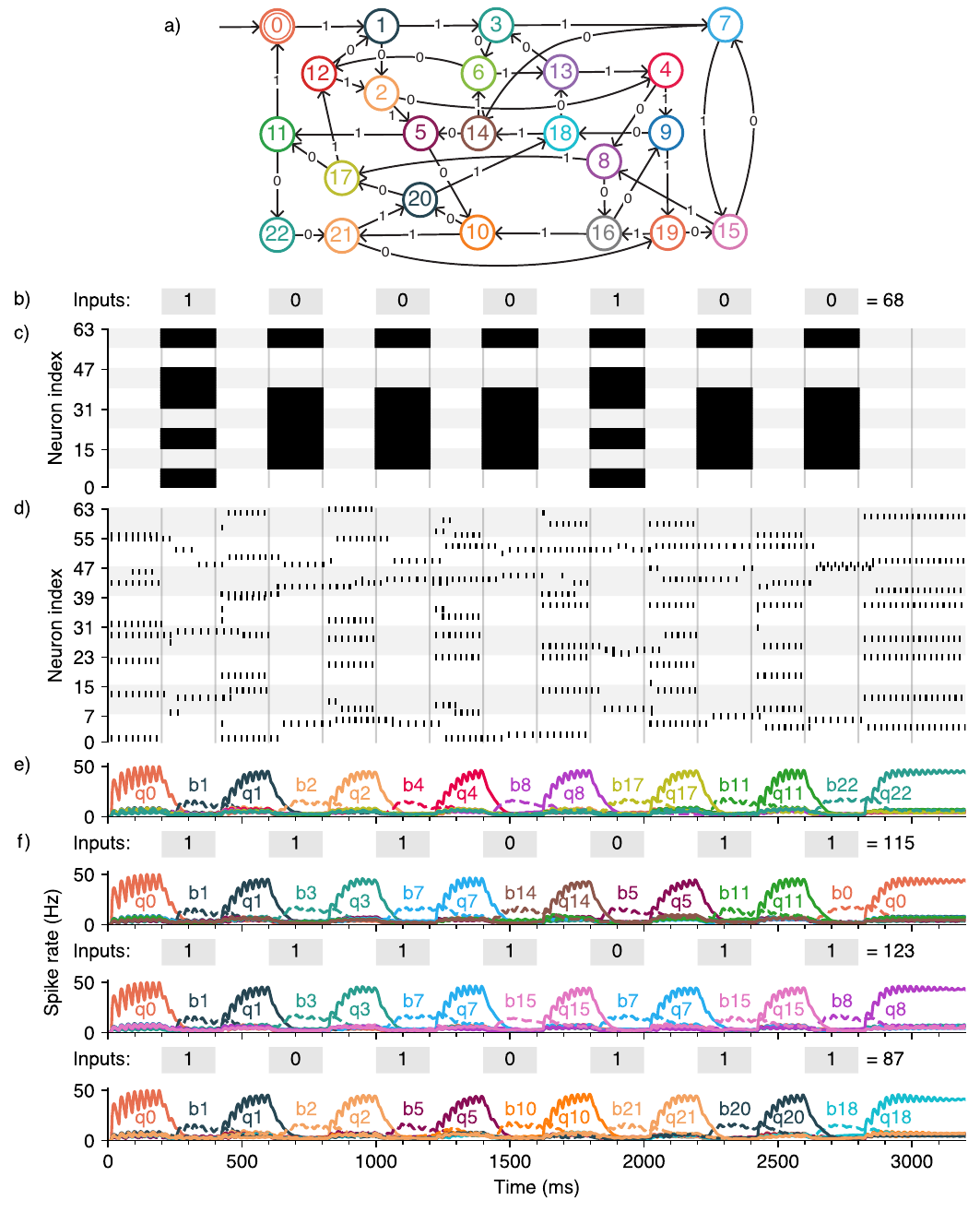}
\caption{An RSNN performing a walk on a 23-state DFA, using noisy 1-bit weights. \textbf{a)} The embedded DFA. If a binary number is input most-significant-bit first, the final state indicates the result of the input $\mathrm{mod} \: \: 23$. \textbf{b)} The symbolic input to the RSNN. \textbf{c)} The input vector to the RSNN at any time, which each masks out half of the neurons therein. Each different input $s$ corresponds to a different hypervector $\vec{s}$. Only the first 64 neurons are shown. \textbf{d)} A spike raster plot of activity within the RSNN. When the input to the network is constant, the network stabilises in an attractor state. When the input changes, some blocks are masked out or revealed, which may cause transitions to a new attractor state. The block-WTA mechanism ensures that only one neuron is persistently active in each block. \textbf{e)} The kernel-filtered mean firing rate of the neurons active in the attractor states $\vec{q}_0, \ldots , \vec{q}_{22}$, as well as the transition-facilitating bridge states $\vec{b}_0, \ldots , \vec{b}_{22}$.
The bridge states are represented by dashed lines and coloured the same as their corresponding $q$ attractor states.
% Interpreting the sequence of inputs given as a binary number, the final inhabited attractor state indicates the result of the number modulo 23.
We here serially input the number 68 in binary format, and the RSNN correctly halts in the $q_{22}$ state. \textbf{f)} The same RSNN is given different sequences of inputs, and the RSNN performs the correct walk between attractor states in all cases.}
\label{fig:sim_walk}
\end{figure*}

The simulated RSNN consists of $N$ leaky integrate-and-fire (LIF) neurons, with all-to-all synaptic connectivity, except between neurons in the same block, which have \mbox{winner-take-all (WTA)} connectivity between themselves (see Methods). This enforces that only one neuron in each block may spike at once, thereby restricting the neural activity to have a sparse block code structure.

\begin{figure}
\centering
\includegraphics[width=0.9\linewidth]{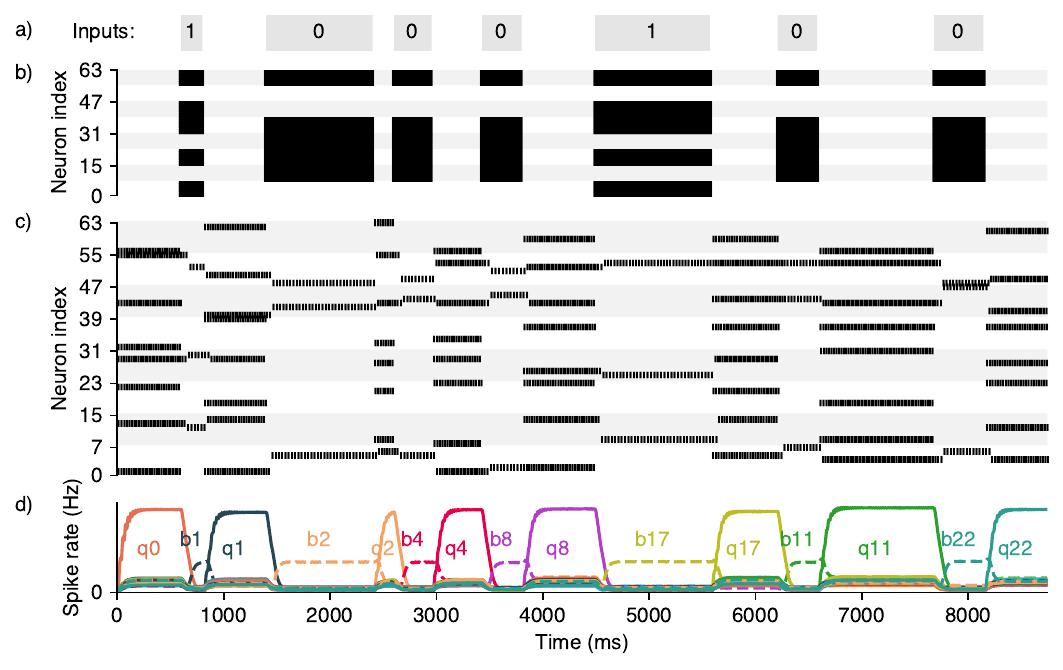}
\caption{The simulated RSNN with an irregular input timing scheme. \textbf{a)} The symbolic input to the network. Inputs were given for between \qty{200}{ms} and \qty{1000}{ms}. \textbf{b)} The corresponding input hypervector masking the neurons. \textbf{c)} A raster plot of spike activity in the RSNN. \textbf{d)} The mean firing rate of the neurons in each attractor state. The network is not reliant on an exact input timing scheme to perform the correct sequence of transitions. There is however still a minimum duration for which an input (or lack thereof) must be given, determined by the synaptic time constant.}
\label{fig:sim_inhomo_timing}
\end{figure}

Although the method generalises to arbitrary DFAs~\cite{cotteret_vector_2024}, we chose to implement a 23-state DFA which can perform modular division of binary numbers \mbox{(Fig.~\ref{fig:sim_walk})}, as it is a clear example of useful computation. Two inputs, $s_0$ and $s_1$, with associated random hypervectors $\vec{s}_0$ and $\vec{s}_1$, are input to the network to represent ``0'' and ``1'', respectively. A sequence of inputs thus corresponds to a dividend $d$ in binary representation, and the final state corresponds to the result of the modular division $d \ \mathrm{mod} \ 23$. The full state transition function is described by the transitions $q_n \overset{s_0}{\longrightarrow} q_{(2n \ \mathrm{mod} \ 23)}$ and $q_n \overset{s_1}{\longrightarrow} q_{(2n +1 \ \mathrm{mod} \ 23)}$ for all states $q_n$, $n \in \{0, \ldots, 22 \}$. The weight matrix to implement this DFA was constructed as described in Methods. We then stochastically binarized the ideal weights and added independently-sampled Gaussian noise of standard deviation 0.5 to each binary weight value, such that the two weight distributions were highly overlapping. This demonstrates robustness to low-precision and noisy weights, inherent in mixed-signal neuromorphic hardware platforms. Despite these nonidealities, the RSNN performs the correct walk between attractor states for every input sequence \mbox{(Fig.~\ref{fig:sim_walk})}. The network is not reliant on a specific input timing scheme, as long as the inputs are given for long enough for a transition to take place (Fig.~\ref{fig:sim_inhomo_timing}).
A similar 300-state DFA with ideal weights \mbox{(Fig.~\ref{fig:sim_huge_dfa_results})} was constructed in the same way, to demonstrate the seamless scaling to large state machines.

\subsection{Memristive crossbar hardware implementation}

\begin{figure*}
    \centering
    \includegraphics[width=0.7\linewidth]{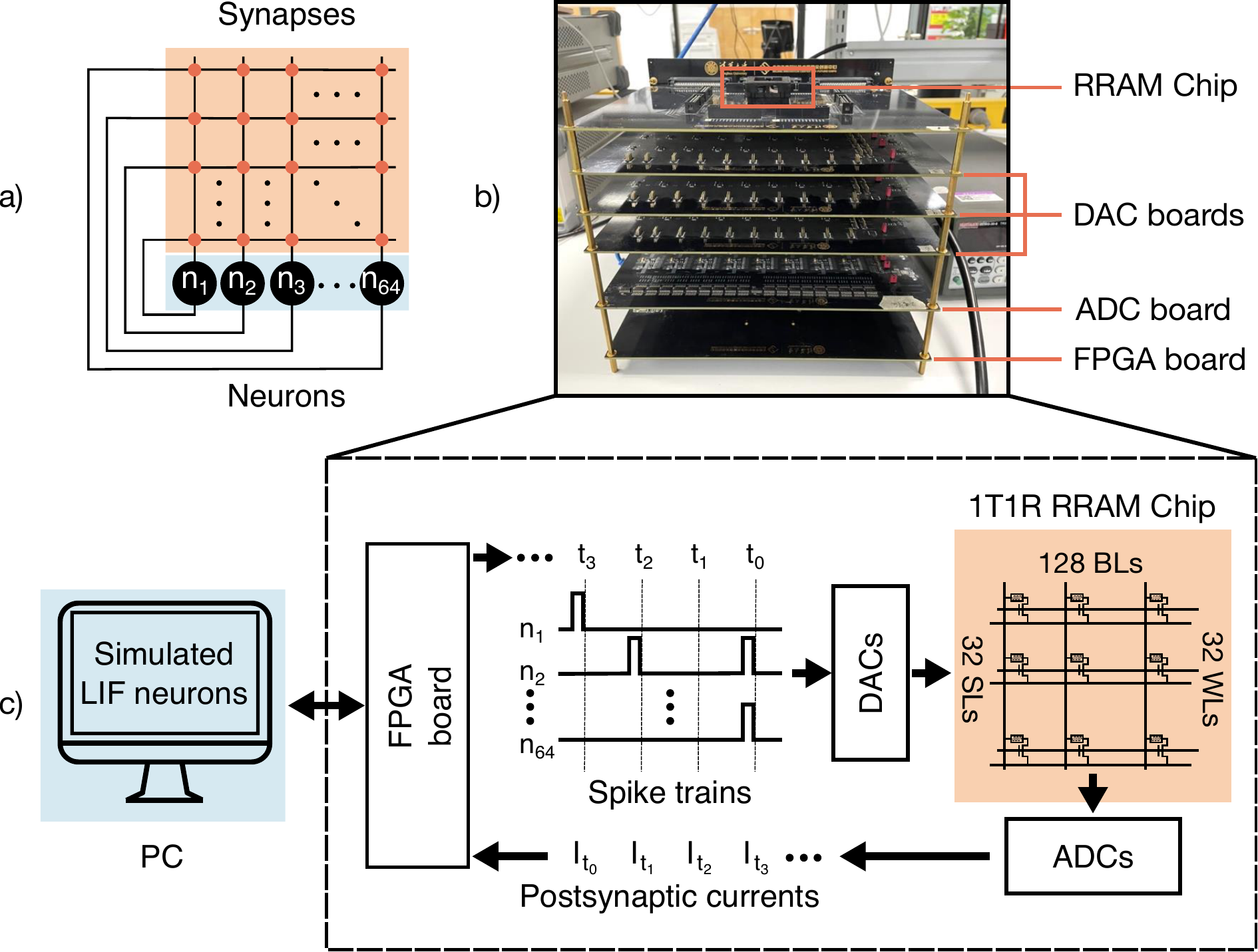}
    \caption{The closed-loop experimental setup for running the RSNN using the 4K RRAM system. \textbf{a)} A simplified schematic of the 64-neuron RSNN. \textbf{b)} The physical RRAM measurement system, containing an FPGA board, three DAC boards, and one ADC board. \textbf{c)} The scheme of running the closed-loop experiment in which the LIF neurons are simulated on a PC. The FPGA board receives the control commands from the PC and operates all word-, source- and bit lines (WLs, SLs, BLs) of the RRAM chip in parallel. At each time step ($t_n$), the spikes generated by the neurons are sent to the synapse array on WLs, which are represented by voltage pulses. Meanwhile, a \qty{0.2}{V} read voltage is applied to the SLs. The postsynaptic currents ($I_{t_n}$) produced by the RRAM cells accumulate on each BL, and the readout results are returned to the PC through the ADCs and FPGA.}
    \label{fig:RRAM_system}
\end{figure*}

To demonstrate that the approach is robust to nonidealities introduced by memristive devices,
we run a closed-loop experiment with synapses on a memristive crossbar and neurons in simulation (Fig.~\ref{fig:RRAM_system}). Whenever the simulated neurons spiked, the memristive crossbar was read by applying the binary vector of spiking neurons as a voltage input vector to the crossbar via the on-board DACs. The readout currents were read by the ADCs and then fed back into the simulation as postsynaptic currents.
With this hybrid ``computer in the loop'' approach, we demonstrate the readiness of the algorithm for implementation on emerging hybrid SNN-memristor platforms~\cite{greatorex_texel_2024}.

\begin{figure*}
    \centering
    \includegraphics[width=0.9\linewidth]{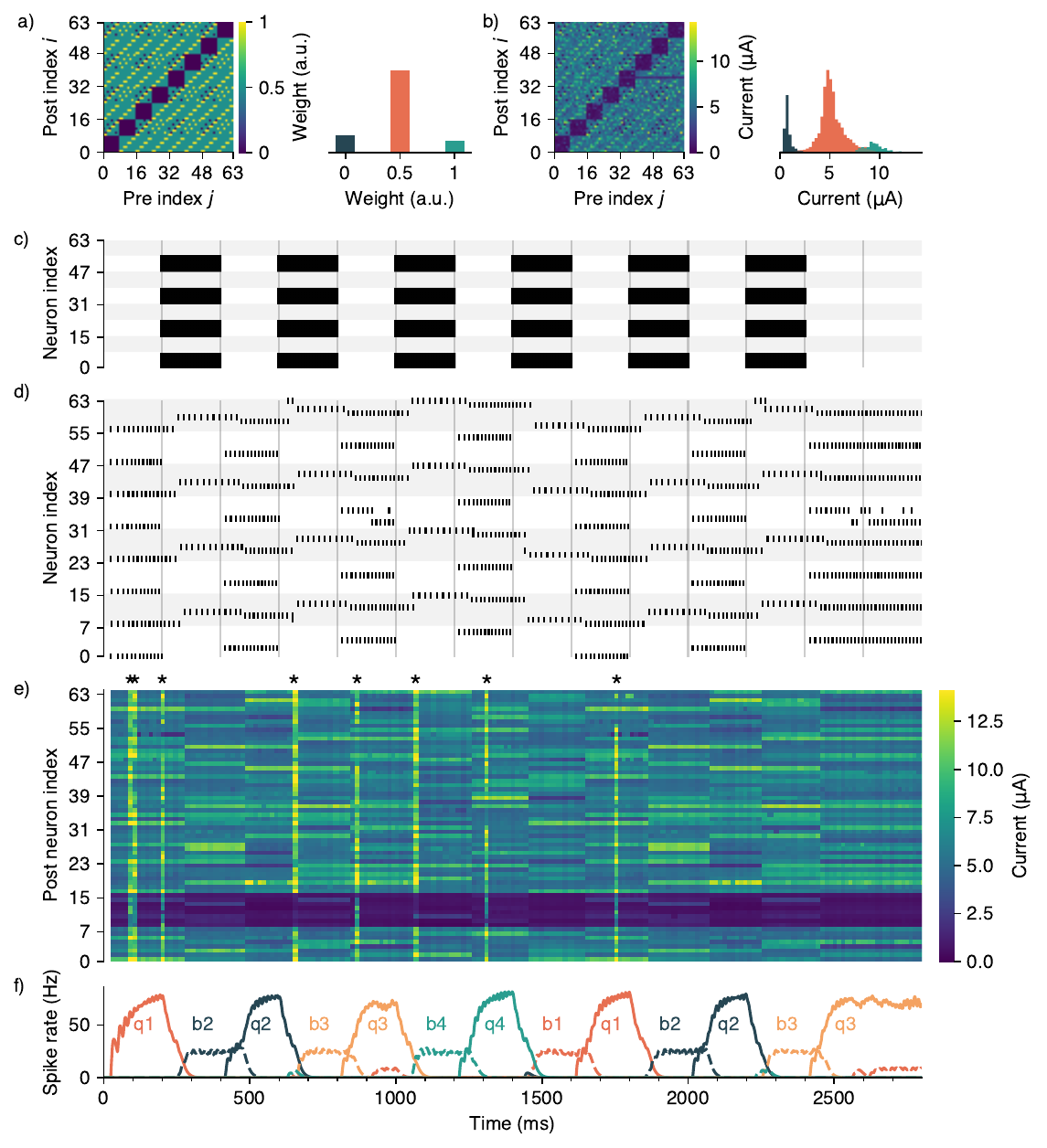}
    \caption{A 4-state DFA embedded into an RSNN using a memristive crossbar with 64$\times$64 RRAM devices as the synaptic weights. The DFA is described by $q_0 \overset{s}{\rightarrow} q_1 \overset{s}{\rightarrow} q_2 \overset{s}{\rightarrow} q_3 \overset{s}{\rightarrow} q_0$ for a single input $s$. \textbf{a)} The ternary weight matrix to be written to the RRAM crossbar, and a histogram of the values. \textbf{b)} Readout currents from each of the 64$\times$64 RRAM devices after programming, and separate histograms for each of the ternary weight values. There is notable mismatch between the measured currents and the ideal values; a row of faulty devices giving almost no current; and devices giving anomalously large currents. \textbf{c)} The masking input to the network. \textbf{d)} A spike raster plot of the neurons within the RSNN. Due to the size constraints introduced by the crossbar, we chose the attractor hypervectors $\vec{q}$ to be orthogonal rather than random. \textbf{e)} Measured postsynaptic current readings from whenever a neuron in the second block spiked, chosen for the prominence of trial-to-trial current variation. The weights between neurons in the same block were programmed to the lowest weight, hence the horizontal band of low currents. At some times, multiple neurons fired within the same time step (labelled by $\ast$). \textbf{f)} The mean firing rates of the neurons in each attractor state. Despite the considerable nonidealities present, the RSNN performs the correct walk between attractor states.}
    \label{fig:dfa_memristor}
\end{figure*}

We used an RRAM crossbar with 4096 devices. Due to this size constraint, we implemented smaller DFAs than in simulation. The weights were mapped to ternary values by thresholding, and then written to three conductance states on the RRAM devices. Although the device programming procedure in principle supports more weight states -- limited by the precision of the ADCs (12-bit) -- relaxation effects in the filamentary RRAM cause the conductance values to disperse in the milliseconds to seconds after programming, reducing the effective precision of the weights~\cite{rram_relaxation_2015, rram_relaxation_2016}. After this dispersion, the RRAM devices store the synaptic weights in a non-volatile fashion~\cite{chen_scaling_2024}. We first implemented a DFA representing a 4-state counter. The RSNN performs the correct walk between attractor states, despite the fixed-pattern-noise and trial-to-trial weight nonidealities in the crossbar current readouts \mbox{(Fig.~\ref{fig:dfa_memristor})}. We then implemented another 4-state DFA but with two inputs, following an identical procedure. For every input sequence tested, the network performs the correct walk between attractor states \mbox{(Fig.~\ref{fig:xbar_bigger})}.

\subsection{Large state machines on digital neuromorphic hardware}

To demonstrate that the approach scales to large state machines on neuromorphic hardware and can be easily transferred to different types of neuromorphic systems, we implemented the RSNN on Intel's neuromorphic research chip Loihi 2. We implemented a functionally equivalent block-WTA mechanism with shunting inhibition synapses using a custom neuron microcode. This is required to give each neuron both a conventional integrative synapse and a shunting inhibition synapse.
When a shunting synapse is stimulated, it resets the membrane potential of the target neuron to a subthreshold value \mbox{(Fig.~\ref{fig:loihi_wta})}, implementing the within-block WTA and the block-wise masking of the RSNN by input.
The integrative synapses implement the recurrent weight matrix as specified in Methods.
We implemented the same 23-state DFA as in simulation \mbox{(Fig.~\ref{fig:sim_walk})}, now quantizing the ideal weights to the available 8-bit values.
\begin{figure*}
\centering
\includegraphics[width=0.9\linewidth]{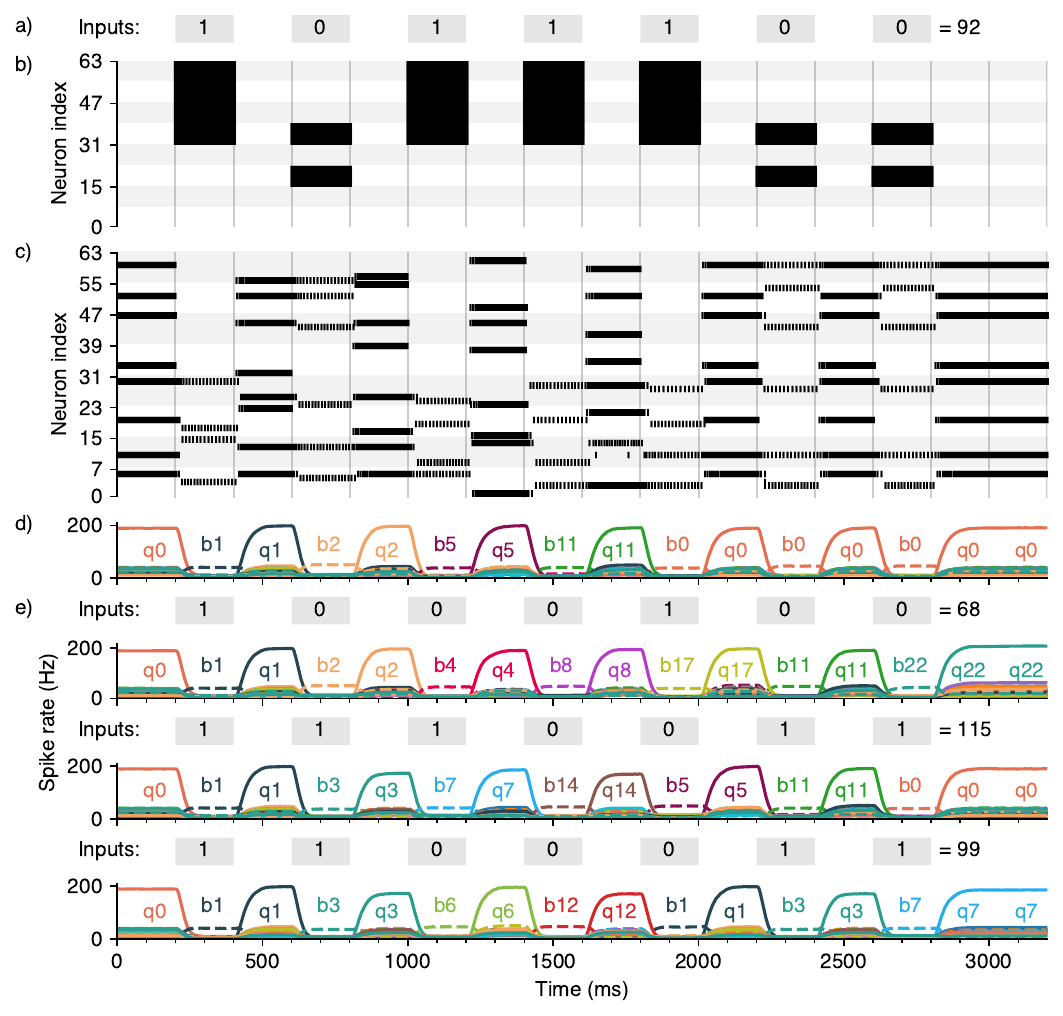}
\caption{
The 23-state DFA on Intel's asynchronous digital neuromorphic research chip, Loihi 2. \textbf{a)} The input symbol to the network at any time, and \textbf{b)} the corresponding input vectors, which mask the network activity. \textbf{c)} A spike raster plot of the first 64 neurons. The shunting-inhibition WTA mechanism ensures that only one neuron in every block may spike at once (see Methods). \textbf{d)} Kernel firing-rate estimates of each $q$ and $b$ state, choosing arbitrarily that 1 time step on Loihi represents \qty{1}{ms}. The sequence of inputs given corresponds to the binary representation of the number 92. The RSNN halts in the $q_0$ state, indicating correctly that 92 is divisible by 23. \textbf{e)} For all sequences of inputs, the correct walk between attractor states is performed.}
\label{fig:raster_walk}
\end{figure*}
The network performs the correct walk between attractor states reliably for all given input sequences \mbox{(Fig.~\ref{fig:raster_walk})}.

\subsection{Network capacity}
\label{sec:network_capacity}

\begin{figure}
\centering
\includegraphics[width=3.7in]{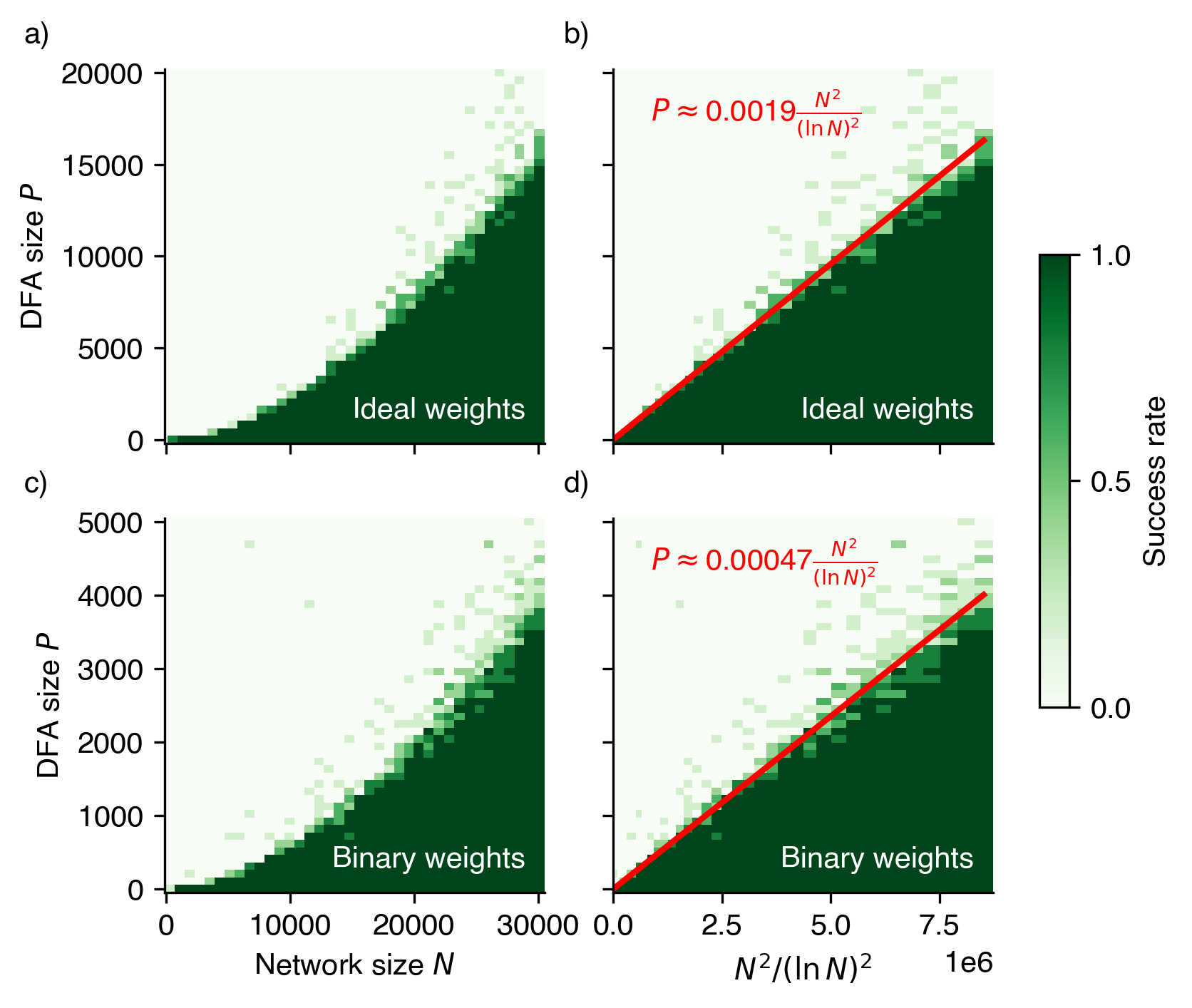}
\caption{Network capacity for varying network size $N$, in terms of the number of DFA states $P$ that can reliably be emulated. Each DFA state has two outgoing edges, like in Fig.~\ref{fig:sim_walk}. \textbf{a)} A network with ideal weights displays an approximately quadratic capacity relation. \textbf{b)} The same data, but plotted against $N^2 / (\log N)^2$, with a separation boundary obtained by a linear SVM.
\textbf{c,d}) A network with stochastic binary weights displays the same scaling relation but reduced by a factor of $\approx 4$. In both cases, the block length $L$ was varied such that $L \propto N / \log N$, with reference values $L_0 = 8$, $N_0 = 2048$.}
\label{fig:sm_capacity}
\end{figure}

Since the RSNNs used in this work are based upon a single-layer attractor network structure, we are limited by the theoretical memory capacity of such networks. That is, for a given network size $N$ and block length $L$ (determining the activation sparsity $1/L$), a maximum number of attractors can be embedded before unwanted cross-talk between patterns leads to a breakdown in attractor dynamics \cite{amit_modeling_1989}. This limits the maximum size of DFA that can be embedded for a fixed $N$. For dense Hopfield networks, this results in the well-known linear capacity limit of $P < 0.14N$, where $P$ is the number of storable patterns \cite{hopfield_neural_1982}. For sparse attractor networks, the capacity instead scales almost quadratically as $P \propto N^2 / (\log N)^2$, provided the sparsity also scales appropriately \cite{amari_characteristics_1989, tsodyks_enhanced_1988, knoblauch_memory_2010}. We conducted non-spiking simulations to numerically investigate whether these scaling relations also hold with our approach, for both real ideal and stochastically binarised weights. We found that the capacity is proportional to $N^2 / (\log N)^2$ in both cases, while the network capacity in the binary weight case is approximately 4$\times$ smaller than with ideal weights (Fig.~\ref{fig:sm_capacity}).

\section{Discussion}

It is widely believed that recurrent attractor dynamics are fundamental to the brain's ability to temporarily represent information in stable neuron firing patterns, without the need for components with intrinsically-long timescales \cite{dagostino_denram_2024}, and despite the nonidealities present in biological systems~\cite{little_existence_1974, rolls_mechanisms_2013, chaudhuri_computational_2016, schneidman_weak_2006, khona_attractor_2022}.
Networks that switch between discrete attractor states are used to explain neural recordings and are thought to underlie higher decision-making processes~\cite{mante_context-dependent_2013,sussillo_opening_2013,miller_itinerancy_2016, dayan_simple_2008}.
These robust dynamics inspired the embedding of controlled attractor transition dynamics in RNNs, in particular in neuromorphic systems that mimic the brain's parallelism and asynchrony, but suffer from a similar inherent unreliability of single components as biological systems \cite{rutishauser_state-dependent_2009} (see \cite{cotteret_vector_2024} for a more comprehensive overview).

\subsecno{Distributed representations are scalable and robust}

In previous efforts to embed attractor-based state machines into spiking neuromorphic hardware~\cite{neftci_synthesizing_2013, liang_neuromorphic_2019, baumgartner_visual_2020}, each state was represented by a dedicated orthogonal (non-overlapping) population of neurons with a synaptic gating mechanism to trigger state transitions.
Although this simplifies the state encoding, these localist state representations limit the robustness, flexibility, and scalability of the network in comparison to a distributed approach~\cite{rumelhart_parallel_1986}. If each state is represented by a unique population of $M$ neurons that are active for only one pattern, then a network of $N$ neurons can represent at most $N/M$ patterns.
In contrast, sparse distributed single-layer attractor networks can store $N^2 / (\log{N})^2$ patterns, making more efficient use of the available synaptic resources~\cite{amari_characteristics_1989, tsodyks_enhanced_1988, knoblauch_memory_2010}.
Additionally, adding new states with localist representations requires adding or finding new neurons that are inactive for every other pattern. This is at odds with the mixed selectivity of neurons found throughout the brain~\cite{fusi_why_2016}.

We instead utilised the distributed representational framework of VSAs, in which high-dimensional random vectors can be composed to represent arbitrary data structures and algorithms~\cite{kleyko_survey_2022}.
% We instead utilised the distributed representational framework of VSAs, with which it has been shown possible to achieve the approximately quadratic scaling for state machines embedded into RNNs~\cite{cotteret_vector_2024}.
% In VSAs, symbolic data is represented by high-dimensional random vectors which can be composed to represent arbitrary data structures and algorithms~\cite{kleyko_survey_2022}.
The state-representing hypervectors thus have a nonzero probabilistic degree of overlap, but it is nonetheless possible to generate an exponential (in $N$) number of hypervectors for which significant overlap between any pair of hypervectors is unlikely (Section \ref{sec:sbc_capacity})~\cite{thomas_theoretical_2022, clarkson_capacity_2023}. The sparsity of the hypervectors not only increases the network's memory capacity, but makes them inherently suitable for exploiting the efficiency of sparse activity in neuromorphic hardware \cite{boahen_dendrocentric_2022}.

The distributed representations also make the network highly robust to various nonidealities, such as noise, individual component failure and nonideal neuron dynamics. Their robustness to low-precision (quantized) and inaccurate synaptic weights is particularly relevant for memristive in-memory applications.
This has prompted recent interest in combining VSAs with memristive crossbars~\cite{karunaratne_-memory_2020, barkam_reliable_2023} and spiking neural networks~\cite{frady_robust_2019, renner_neuromorphic_2024, zou_memory-inspired_2022, morris_hyperspike_2022}, where synaptic and neural nonidealities are abundant.
The results obtained from simulations and the closed-loop memristive crossbar setup confirm the system's reliable operation under hardware constraints, notably without requiring parameter fine-tuning, while the restriction of binary weights causes only a constant factor decrease in the network capacity.

\subsecno{Simultaneous auto- and heteroassociative memory}

By exploiting the pseudo-orthogonality property of the hypervector binding operation, we were able to superimpose autoassociative and heteroassociative outer-product terms in the recurrent weight matrix without incurring considerable interference between them. Transitions are triggered by masking a subset of the neurons -- constituting an effective unbinding operation -- while in previous work, additional mechanisms were required to modulate between attractor and transition dynamical regimes~\cite{horn_neural_1989, sommer_synfire_2005, kambara_role_1997, gutfreund_processing_1988, amit_neural_1988, drossaers_hopfield_1992, chen_attractor-state_2020, peretto_collective_1986}.
Furthermore, the simplicity of this construction method means that the weight matrix could reasonably be learned via local Hebbian learning rules, such that states and transitions are learned and unlearned as needed to solve a particular task~\cite{osipov_associative_2017}.

\subsecno{Fault-tolerant computation}

Programming a certain DFA into an RSNN in one shot may be sufficient for some applications, however. In space, ionizing solar radiation causes unwanted bit-flips in systems using conventional charge-based memories, and so necessitates fault-tolerant computation schemes. Our system is intrinsically robust to such events since the attractor dynamics quickly reverse any erroneous flips in the neural state. Furthermore, RRAM-based memories are highly resistant to solar radiation, making memristive VSA-programmed attractor-based neuromorphic implementations of conventional algorithms potentially suitable for space and other high-energy applications~\cite{maestro-izquierdo_gamma_2021}.

The flexibility and robustness of the model enable a wide range of applications for neuromorphic hardware. It could be used to coordinate information flow in complex neuromorphic systems~\cite{baumgartner_visual_2020} or be directly incorporated into existing works leveraging VSA representations~\cite{kleyko_survey_2023}, such as online learning of robotics schemas~\cite{neubert_introduction_2019} or general computation with distributed stack machines~\cite{yerxa_hyperdimensional_2018, kleyko_vector_2022}.

\subsecno{Neuromorphic hardware abstraction}

By utilizing distributed symbolic representations, we programmed an RSNN from an abstract higher level of description, which is separated from the exact details of the hardware. It relied on the existence of a fixed stereotypical low-level neural motif (the block-WTA connectivity), but adjusted only the connections between neurons in different blocks, to easily program the same neural hardware for different purposes.
This ability to abstract the function of lower layers and then build upon them relatively carefree is the foundation of conventional digital design approaches and bears resemblance to Marr's levels of analysis~\cite{marr_vision_1982, kleyko_vector_2022}.

The distributed representations of VSAs are especially suited as an abstraction layer to divorce high-level function from low-level neural dynamics, in part due to their invariance of the choice of hypervector representation~\cite{kleyko_vector_2022, kleyko_survey_2022}. For example, in place of sparse block code representations, we could have instead used phasor representations \cite{noest_phasor_1987, plate_holographic_1995} or their sparse variants, which have an elegant connection to the periodic spiking of resonate-and-fire neurons~\cite{frady_robust_2019,renner_neuromorphic_2024}. The same algorithm could be implemented on different neuromorphic hardware platforms, using whichever representation is best suited for the available neuron circuit.
Neuromorphic computing would greatly benefit from a high-level neurally-inspired abstract programming language, which can be robustly implemented across different neuromorphic hardware platforms~\cite{schuman_opportunities_2022, kleyko_vector_2022, jaeger_toward_2023}.
Although there have been many works towards such a unification~\cite{stefanini_pyncs_2014, davison_pynn_2009, aimone_composing_2019, zhang_system_2020, eliasmith_unified_2005, pedersen_neuromorphic_2024}, the majority of these focus on finding common ways to interface with different neuromorphic hardware platforms or to compose functional SNN modules, rather than on defining higher-level abstract primitives that are invariant of the neural representations on the hardware on which they are embedded.

This work demonstrates the capability of distributed representations for embedding computation through neural dynamics into neuromorphic hardware~\cite{kleyko_vector_2022}.
We embedded arbitrary state machines into RSNNs, as they are a clear example of multi-timescale computation: the switching between states is fast, but the autoassociative attractor dynamics ensure stability on long timescales~\cite{hopfield_neural_1982, amit_modeling_1989}. The extension of this approach to embedding general computational primitives beyond DFAs into neuromorphic RSNNs, such as continuous attractor networks \cite{khona_attractor_2022}, presents an exciting direction for future research.
The fully-distributed representations ensure increasing robustness with increasing dimensionality, while the invariance to the choice of hypervector representation -- many of which have elegant spiking implementations -- may allow neuromorphic hardware to be programmed at a level that is independent of the underlying neural representations, but that can reliably and robustly be implemented on different neuromorphic hardware platforms without the need for fine-tuning.

\section{Methods}
\label{sec:methods}

\subsection{VSA arithmetic}

% % \color{blue}
VSAs allow compositional data structures to be represented in a fully distributed manner, using high-dimensional random vectors, hypervectors, as the smallest units of representation \cite{kleyko_survey_2022, gayler_multiplicative_1998, kanerva_hyperdimensional_2009, plate_holographic_1995}. They rely upon the fact that independently-generated hypervectors will very likely be approximately orthogonal to each other (the \textit{pseudo-orthogonality} property) and so new hypervectors can easily be generated to represent new items. For most hypervector representations, the similarity between hypervectors is defined by the normalised inner product
\begin{equation}
    \mathrm{sim}(\vec{a}, \vec{b}) = \frac{\vec{a} \cdot \vec{b}}{\abs*{\vec{a}}_2 \abs*{\vec{b}}_2 }
\end{equation}
such that, for a set of independently-generated hypervectors $\{\vec{x}_\nu\}$ the pseudo-orthogonality property is then written as
\begin{equation}
    \mathrm{sim}(\vec{x}_\nu , \vec{x}_\mu) \begin{cases}
        = 1 \quad \text{if} \quad \mu = \nu \\
        \approx 0 \quad \text{otherwise}
    \end{cases}
\end{equation}

Each VSA model is equipped with two operators to generate hypervectors which represent relations between other hypervectors, and thus also the symbols they represent. They are the superposition and binding operators, often corresponding simply to component-wise addition and multiplication respectively. The superposition operator produces a hypervector of the same dimensionality, which retains similarity to its operands. That is, if $\vec{a}$ and $\vec{b}$ are independent hypervectors, then
$
\mathrm{sim}(\vec{a} \oplus \vec{b}, \vec{a}) \approx \mathrm{sim}(\vec{a} \oplus \vec{b}, \vec{b}) \approx 1
$
where ``$\oplus$'' represents a general superposition operator. Superpositions of hypervectors are thus used to efficiently represent sets of objects \cite{thomas_theoretical_2022, clarkson_capacity_2023}.
 
The binding operator produces a hypervector that is dissimilar to both its operands. That is, if $\vec{a}$ and $\vec{b}$ are again independently-generated, then
$
    \mathrm{sim}( \vec{a} \odot \vec{b} , \vec{a}) \approx \mathrm{sim}( \vec{a} \odot \vec{b} , \vec{b}) \approx 0
$
where ``$\odot$'' is a general binding operation. Binding may be inverted by unbinding the same hypervector, i.e. $(\vec{a} \odot \vec{b}) \odot^{-1} \vec{b} = \vec{a}$. Superimposing pairs of bound hypervectors could then be used to create a hypervector representing a set of key-value pairs \cite{kanerva_hyperdimensional_2009}, for example, but more complex objects can similarly be represented such as graphs \cite{poduval_graphd_2022, kleyko_survey_2022}, algorithms \cite{yerxa_hyperdimensional_2018} or images \cite{frady_computing_2022}.

In this work, we used two types of VSA models: dense bipolar hypervectors, also known as Multiply-Add-Permute (MAP) hypervectors, and Sparse Block Code (SBC) hypervectors \cite{kleyko_survey_2022}. MAP hypervectors are randomly sampled from $\{ -1, 1\}^N$, with approximately equal numbers of positive and negative entries. The binding operation for MAP hypervectors is the Hadamard product ``$\had$'' (element-wise multiplication), which also serves as the inverse binding operation \cite{gayler_multiplicative_1998}. SBC hypervectors are binary-valued and have a sparse-block structure. That is, the $N$ components are divided into $M$ equally-sized blocks of length $L = N / M$, and we randomly choose exactly one component in each block to be 1-valued, with the rest 0 \cite{laiho_high-dimensional_2015,frady_variable_2023}. 
By binding MAP and SBC hypervectors with a Hadamard product operation, we produce a bipolar sparse hypervector that is pseudo-orthogonal to the original SBC hypervector (see Section \ref{sec:sm_sbc_map} for further discussion on the joint MAP-SBC VSA model). We can then superimpose weight matrices constructed using these hypervectors, each corresponding to a different network behaviour, without significant interference. Masking out some components constitutes an effective Hadamard unbinding operation.  This operation allows that terms in the weight matrix bound to a MAP hypervector (and thus pseudo-orthogonal to all SBC hypervectors) may be unbound and align with the network state. As a result,  state transitions in the RSNN can be triggered by selectively masking neurons with inhibitory input.

% \color{black}

\subsection{State machine embedding}
\label{sec:W_construction}

A DFA is a finite state machine consisting of a set of states $Q$, an initial state $q_0 \in Q$, a finite set of input symbols $S$, and a state transition function $F: Q \times S \rightarrow Q$. In addition, a DFA denotes a subset $A \subseteq Q$ of its states as ``accepting'' states. A string of input symbols is then said to have been accepted by the DFA if it terminates in one of the acceptance states. As such, DFAs can be used to perform pattern matching -- accepting or rejecting strings based on a predefined set of rules, regular expressions~\cite{minsky_computation_1967}.

The DFA is embedded into the RSNN by first generating for each state $q \in Q$ an SBC hypervector $\vec{q} \in \{0,1\}^N$ to represent it.
For every input symbol $s \in S$, a dense binary hypervector $\vec{s} \in \{0, 1\}^N$ is generated, which, when input to the network, should trigger the correct transitions between attractor states. We restrict these hypervectors to be block-constant, with the same block length $L$ as the SBC hypervectors, such that components in the same block have the same value. The 0 and 1 entries were chosen with equal probability. We will denote $\bar{\vec{s}} \in \{-1,1\}^N$ as the equivalent bipolar representation of these hypervectors, given by $\bar{\vec{s}} = 2 \vec{s} -1$. For every state $q$, an additional SBC vector $\vec{b}$ is generated, which we refer to as its \textit{bridge} state. They will be attractors of the RSNN dynamics only when there is input to the network. Transitions will be constructed such that, when we give input to the RSNN, it will not transition directly from a source state $\vec{q}$ to target state $\vec{q}^\prime$, but rather will first transition from $\vec{q}$ to $\vec{b}^\prime$, the bridge state belonging to the target state $\vec{q}^\prime$. When the input is removed, the RSNN will finally flow from the $\vec{b}^\prime$ to $\vec{q}^\prime$ state, completing the transition. Splitting each transition up in this way ensures they are not dependent upon a specific input timing scheme \mbox{(Fig.~\ref{fig:sim_inhomo_timing})}. The construction of the recurrent weight matrix $\mat{W} \in R^{N \times N}$ can be divided into three parts, first to embed the $\vec{q}$ states as attractors
\begin{equation}
\begin{split}
    \mat{W}_{\text{attr.}}  = \sum_{q \in Q}  (\vec{q}-f)(\vec{q}-f)\T
\end{split}
\label{eqn:W_attr}
\end{equation}
where the coding level $f = 1/L$ is the fraction of nonzero elements in each $\vec{q}$ hypervector \cite{tsodyks_enhanced_1988, amari_characteristics_1989}. We then embed the bridge states as
\begin{equation}
% \begin{split}
    \mat{W}_{\text{brdg.}}  = \sum_{q \in Q}  \Big[  (\vec{q}-f)(\vec{b}-f)\T 
     + \sum_{s \in S} (\vec{b} - \vec{q}) \big( (\vec{b} -f) \had \bar{\vec{s}} \big)\T \Big]
% \end{split}
\label{eqn:W_bridge}
\end{equation}
where ``$\had$'' is the Hadamard product binding operation.
We are thus using an unorthodox mixture of the MAP~\cite{gayler_multiplicative_1998, kleyko_survey_2022} and SBC~\cite{laiho_high-dimensional_2015, frady_variable_2023} VSA models (Section \ref{sec:sm_sbc_map}).
These terms cause the bridge states $\vec{b}$ flow to their corresponding $\vec{q}$ states, except when a valid input is present, then they themselves become attractors of the RSNN's dynamics. To store transitions between states, outer product terms are then similarly added for each non-loop transition, the set of which we define as $ E =  \{ (q, s, q^\prime ) \: \ \text{s.t.} \ \: q^\prime = F(q,s), \: q^\prime \neq q \}$. They are included as
\begin{equation}
\begin{split}
    \mat{W}_{\text{trans.}}  = \sum_{\eta \in E}  (\vec{b}^\prime - \vec{q}) \big( (\vec{q} - f ) \had \bar{\vec{s}} \big)\T
\end{split}
\label{eqn:W_tran}
\end{equation}
where $\vec{q}$ is the attractor state for source state $q$ for edge $\eta$, and $\vec{b}^\prime$ the bridge state belonging to the target state $q^\prime = F(q, s)$, as defined by $E$. The full weight matrix is then given by the superposition $\mat{W} = \mat{W}_\text{attr.} + \mat{W}_\text{brdg.} + \mat{W}_\text{trans.}$.

\subsection{RSNN attractor dynamics}
\label{sec:rsnn_dynamics}

To understand how the described weight construction method results in the desired DFA being emulated, we coarsely model the RSNN by a neural state vector $\vec{z} \in \{0, 1\}^N$, indicating which neuron is active. We then consider the postsynaptic sum $\vec{h} = \mat{W}\vec{z}$, with and without input to the RSNN.
It is instructive to work with the expectation of $\vec{h}$, where terms containing inner products between pseudo-orthogonal hypervectors can be discarded. We must, however, keep in mind that for finite $N$, the postsynaptic sum will only approximate its expected value, due to the accumulation of nonzero inner products between realised hypervectors. See~\cite{thomas_theoretical_2022, clarkson_capacity_2023} for a rigorous theoretical analysis. We first consider the dynamics when the network is in a DFA state, $\vec{z} = \vec{q}^\mu$. The expected postsynaptic sum $\langle \vec{h} \rangle$ is then given by
\begin{equation}
\begin{split}
\langle \vec{h} \rangle & = \bigl \langle \mat{W} \vec{q}^\mu\bigr \rangle \\
& =  \bigl \langle \mat{W}_\text{attr.} \vec{q}^\mu\bigr \rangle + \bigl \langle \mat{W}_\text{brdg.} \vec{q}^\mu\bigr \rangle+ \bigl \langle \mat{W}_\text{trans.} \vec{q}^\mu\bigr \rangle  \\
& \propto (\vec{q}^\mu - f) + \vec{0} + \vec{0} \\
& = \vec{q}^\mu - f
\end{split}
\label{eqn:mean_h_attr}
\end{equation}
such that the state $\vec{q}^\mu$ projects back to itself (Section \ref{sec:h_calc}). Neither $\mat{W}_\text{brdge.}$ nor $\mat{W}_\text{trans.}$ contribute to $\langle \vec{h} \rangle$, since they are constructed either from independent SBC hypervectors, or the same SBC hypervector $\vec{q}^\mu$ but bound to a bipolar hypervector $\overline{\vec{s}}$, both of which are pseudo-orthogonal to the state $\vec{q}^\mu$.
Hence, in the large-$N$ limit, and sufficiently close to the states $\vec{q}$, we can approximate the dynamics as being governed only by the $\mat{W}_\text{attr.}$ matrix. Since $\mat{W}_\text{attr.}$ is symmetric, this results in fixed-point attractor dynamics with an energy landscape consisting of stable minima at the states $\vec{q}$ (Section \ref{sec:energy}) ~\cite{tsodyks_enhanced_1988, amari_characteristics_1989, cohen_absolute_1983}.

\subsection{RSNN transition dynamics}
\label{sec:rsnn_transition_dynamics}

We now consider how $\langle \vec{h} \rangle$ changes when we apply an input $s^\lambda \in S$ to the RSNN, which we assume corresponds to a valid transition $\eta = (q^\mu, s^\lambda, q^{\mu \prime}) \in E$ from state $q^\mu$ to $q^{\mu \prime}$. Inputting $s^\lambda$ means applying the corresponding hypervector $\vec{s}^\lambda \in \{0, 1\}^N$ as a mask to the RSNN, such that all neurons in the blocks for which $\vec{s}^\lambda$ is zero are prevented from spiking.
The expected postsynaptic sum is then given by
\begin{equation}
\begin{split}
\langle \vec{h} \rangle & = \bigl \langle \mat{W} ( \vec{q}^\mu \land \vec{s}^\lambda )\bigr \rangle \\
& =  \bigl \langle \mat{W}_\text{attr.}(  \vec{q}^\mu \land \vec{s}^\lambda ) \bigr \rangle  + \bigl \langle \mat{W}_\text{brdg.}(  \vec{q}^\mu \land \vec{s}^\lambda ) \bigr \rangle +  \bigl \langle \mat{W}_\text{trans.}(  \vec{q}^\mu \land \vec{s}^\lambda ) \bigr \rangle  \\
& \propto (\vec{q}^\mu - f) +\vec{0} +  (\vec{b}^{\mu \prime} - \vec{q}^\mu) \\
& = \vec{b}^{\mu \prime} - f
\end{split}
\label{eqn:mean_h_tran}
\end{equation}
where $\land$ is a component-wise $\mathrm{AND}$, implementing the masking operation (Section \ref{sec:h_calc}). Since the masked $\vec{q}^\mu$ state now does not project back to itself, but to $\vec{b}^{\mu \prime}$ (the bridge state belonging to the target $q^{\mu\prime}$ state), the RSNN transitions to the $\vec{b}^{\mu \prime}$ state.
By applying $\vec{s}^\lambda$ as a mask to the RSNN, we caused heteroassociative terms to be projected out of the edge summation in $\mat{W}_\text{trans.}$, which push the RSNN from $\vec{q}^\mu$ to $\vec{b}^{\mu \prime}$.

A similar calculation can be done to show that the bridge states $\vec{b}$ are attractors only when the network is being masked by a valid input, i.e. $ \langle \mat{W} (\vec{b}^{\mu \prime} \land \vec{s} )  \rangle = \vec{b}^{\mu \prime} - f$. When the input is removed, however, this becomes $\langle \mat{W} \vec{b}^{\mu \prime}  \rangle = \vec{q}^{\mu \prime} - f$, driving the RSNN towards the target state, completing the transition. By applying a sequence of inputs to the RSNN as masks (with pauses in between), we cause the RSNN to perform the desired sequence of transitions between attractor states. See Fig.~\ref{fig:sm_transition_mechanism} for a depiction of how this unfolds at the neural level of description. Including the bridge states in the state machine construction ensures that transitions are not dependent upon a specific input timing scheme, since an input must be given -- and crucially then removed -- to complete a transition \mbox{(Fig.~\ref{fig:sim_inhomo_timing})}~\cite{cotteret_vector_2024}. If an input $s$ were given that did not correspond to a valid transition for the current state $q$, no transition occurs.

Constructing attractor networks with transition dynamics by superimposing symmetric autoassociative terms and asymmetric heteroassociative terms, is not a new idea~\cite{horn_neural_1989}. In previous works, to ensure that neither attractor nor transition dynamics dominated always, one had to either ensure a fine balance between the relative strengths of the two terms~\cite{sommer_synfire_2005, kambara_role_1997, buhmann_noise-driven_1987} or to introduce additional mechanisms to modulate their relative strengths, such as delayed synapses~\cite{gutfreund_processing_1988, amit_neural_1988, drossaers_hopfield_1992, kleinfeld_sequential_1986} or short-term synaptic plasticity~\cite{chen_attractor-state_2020, peretto_collective_1986}. By exploiting the properties of VSAs, we have achieved the same result without such mechanisms and without sacrificing robustness. By adding to $\mat{W}$ latent heteroassociative terms which were each bound to a bipolar hypervector $\overline{\vec{s}}$, in the absence of input, we could ignore their contribution to the RSNN dynamics, due to them being pseudo-orthogonal to the network state \mbox{(Equation \ref{eqn:mean_h_attr})}. However, when we mask the RSNN with the same $\vec{s}$ -- an effective unbinding operation -- these heteroassociative terms are revealed, and trigger the transition dynamics \mbox{(Equation \ref{eqn:mean_h_tran})}. Crucially, this is achieved without ever altering the synaptic weight matrix $\mat{W}$ during operation. We use the masking mechanism to realise an unbinding operation because it is not reliant on the synchronous arrival of inputs~\cite{cotteret_vector_2024}.

\subsection{Spiking neuron model}

We simulate leaky integrate-and-fire neurons (LIF), defined by
\begin{equation}
\label{eqn:LIF}
\frac{\de u_i}{\de t} = -\frac{1}{\tau_m}(u_i - u_\text{rest}) + \frac{1}{C} I_i(t)
\end{equation}
where $u_i$ is dynamical membrane voltage of the neuron $i$, $\tau_m$ and $u_\text{rest}$ are the membrane time constant and rest potential respectively, $I_i$ is the postsynaptic current to that neuron, and $C = \qty{1}{F}$. When $u$ exceeds the spiking threshold $u_\theta$, the neuron emits a spike and $u$ is quickly clamped to a subthreshold reset voltage (here \qty{0}{mV}) for an amount of time determined by the refractory period $\tau_{\text{ref}}$.
The postsynaptic currents $I_i(t)$ we modelled as second-order low-pass filters over spike inputs,
\begin{equation}
\label{eqn:alpha_PSP}
\begin{split}
    \tau_\text{syn} \frac{\de I_i}{\de t} &= - I_i + J_i(t) \\
    \tau_\text{syn} \frac{\de J_i}{\de t} & = - J_i + \sum_{j =1}^{N} w_{ij} \sum_{\text{spks }k} \delta(t - t_j^k)
\end{split}
\end{equation}
where we introduce the synaptic time constant $\tau_\text{syn}$, the Dirac delta function $\delta(\cdot)$, and we sum over the spike times $t_j^k$ from all presynaptic neurons $j$, with synaptic weight $w_{ij}$ measured in coulombs.
This has solution
\begin{equation}
\begin{split}
I_i(t) & = \sum_{j=1}^N w_{ij} \Big( K(t) \circledast \sum_k \delta(t - t^k_j) \Big) (t) \\
& = \big[ \mat{W} \vec{z}(t) \big]_i
\end{split}
\label{eqn:sim_current_soln}
\end{equation}
where $K(t)$ is an exponentially-decaying alpha kernel with time constant $\tau_\text{syn}$, $\circledast$ indicates temporal convolution, and $\vec{z}(t)$ defines the kernel-filtered spike activity.
The current is continuous everywhere (in contrast to the more conventional first-order temporal filters~\cite{chicca_neuromorphic_2014}), which ensures that $I_i$ is not affected by the precise timing of presynaptic spikes~\cite{richter_subthreshold_2023}.
In the Loihi implementation, only a first-order synapse was used, so this is not strictly necessary, however.

\begin{figure*}
\centering
\includegraphics[width=0.85\linewidth]{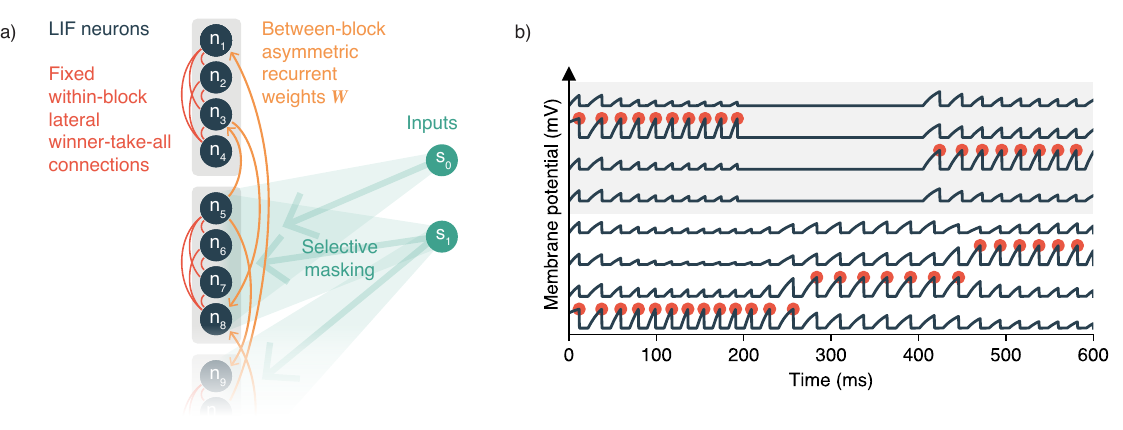}
\caption{\textbf{a)} RSNN network diagram. The neurons are split into blocks of equal length (here $L=4$) with WTA connectivity between all neurons in the same block. The between-block weights $\mat{W}$ are programmed to embed the desired DFA in the RSNN dynamics. The connectivity from each input $s$ is set according to its binary hypervector $\vec{s}$. \textbf{b)} Membrane potential traces $u_i(t)$ of the first 8 neurons within an RSNN. Whenever a neuron spikes (red dots), all neurons within the same block are forced into a refractory state. Thus only the neuron with the greatest postsynaptic current spikes, realising an $\mathrm{argmax}$ function in each block and confining the neural activity to sparse block codes. Between \qty{200}{ms} and \qty{400}{ms}, the second block is masked by an input (either $s_0$ or $s_1$), which causes the network to switch between attractor states.}
\label{fig:diagram_and_membranes_wta}
\end{figure*}

To stabilise the network activity and ensure it always represents an SBC hypervector, we included local winner-take-all (WTA) connectivity between all neurons in the same block, as often done to enforce constraints in RSNNs \cite{Mostafa_etal15}.
It is implemented such that if any neuron spikes, then all neurons in the same block are forced into a refractory state \mbox{(Fig.~\ref{fig:diagram_and_membranes_wta})}. Each block thus performs an $\mathrm{argmax}$ neural activation function on the vector of postsynaptic currents $\vec{I}(t) = \mat{W}\vec{z}(t)$ integrated between spike times. This behaviour is easily implementable in analog VLSI~\cite{cotteret_robust_2023, abrahamsen_time_2004}.
The same mechanism is used to mask out a subset of neurons within the RSNN with external input.
The firing rates $\vec{z}(t)$ and postsynaptic currents $\vec{I}(t)$ approximate the discrete dynamics of the network state $\vec{z}$ and the postsynaptic sum $\vec{h}$ respectively, as described in \mbox{Section \ref{sec:rsnn_dynamics}}.

\subsection{Spiking simulation}
\label{sec:sim_details}

The neuron equations were simulated in the \textit{Brian 2} SNN simulator~\cite{stimberg_brian_2019}, using the Euler ODE solver with a default time step of \qty{0,05}{ms}. The weight matrix to embed the chosen DFA was constructed as described in Methods, with $N = 2048$ neurons and blocks of length $L=8$. These ideal weights were transformed to binary values, to match that biological synapses may, and neuromorphic synapses often do have very few bits of precision \cite{bartol_nanoconnectomic_2015}. This binarisation was performed stochastically, via
\begin{equation}
w_{ij}^\text{binary} = \begin{cases}
    1 \: \: \text{w.p.} \: \: ( 1 + \exp [ - \beta ( w_{ij} - \langle w \rangle ) / \sigma_w ] )^{-1} \\ 0 \: \: \text{otherwise}
\end{cases}
\end{equation}
where $\langle w \rangle $ and $\sigma_w$ are the mean and standard deviation of the ideal weight values respectively, and the steepness parameter $\beta$ is heuristically set to $\beta = 2$. This is a common way to embed analog values into binary-valued memristive devices~\cite{zahari_analogue_2020}. We then add noise to each weight value, to emulate that the distribution of the binary weight states may themselves be imperfect and even overlapping. The nonideal weights used in simulation are then given by
\begin{equation}
w_{ij}^{\text{nonideal}} = | w_{ij}^\text{binary} + \chi_{ij}|
\end{equation}
where $\chi_{ij}$ are independent samples from a centered Gaussian distribution with standard deviation $0.5$. Histograms of the ideal; binary; and nonideal weights are shown in Fig.~\ref{fig:sim_W_distr}. It is well known that attractor networks are robust to such nonidealities and that their effect is to reduce the number of attractors that can be reliably stored~\cite{amit_modeling_1989, sompolinsky_theory_1987, willshaw_non-holographic_1969, knoblauch_memory_2010}.

The RSNN is initialised in the $q_0$ state, and then a sequence of inputs $s_0$ or $s_1$ are given to the network for \qty{200}{ms} at a time, with gaps in-between. We used this particular timing scheme for simplicity, but the proper functioning of the RSNN is not dependent upon the exact timing of inputs. Each input (and gap) may be given for arbitrarily long periods \mbox{(Fig.~\ref{fig:sim_inhomo_timing})}, needing only to be longer than the attractor-switching time, which is of the order $\tau_\text{syn}$.

The firing rate of each attractor state $q$ is given by the dot product of its hypervector $\vec{q}$ and the vector of kernel-filtered spike activity, i.e.
\begin{equation}
\begin{split}
m_q (t) = \frac{1}{M} \sum_{i=1}^N [\vec{q}]_i \big( K(t) \circledast \sum_{\text{spks } k} \delta(t - t_i^k) \big)(t)
\end{split}
\end{equation}
where $m_q(t)$ is the firing rate of state $q$, and $K(t)$ is a normalised alpha kernel with time constant $\tau_\text{readout}$. If the RSNN is exactly representing a state $q$, then $\langle m_q \rangle = \langle \nu \rangle$ where $\langle \nu \rangle$ is the average firing rate of the active neurons. Since the $\vec{q}$ states are not orthogonal, the expected firing rate of the other $q^\prime \neq q$ attractor states $\langle m_{q^\prime} \rangle = \langle \nu \rangle  / L$ is nonzero. The firing rates of the bridge states $\vec{b}$ are likewise calculated.

\subsection{RRAM crossbar experiment}

The neuron equations were simulated in Python using Euler's method with the NumPy package. At each time step, if one or multiple neurons spiked, the memristive crossbar was read, and the measured currents were linearly scaled and then applied to the postsynaptic neurons in simulation.
The experimental measurements were made using non-volatile 1T1R HfO$_x$-based RRAM chips integrating 32$\times$128 devices. The hardware setup is shown in Fig.~\ref{fig:RRAM_system}.
An FPGA board in the measurement system supports the parallel operation of the WLs, BLs, and SLs of the RRAM chip.

With the 4096 RRAM devices, we could create a fully-connected RSNN with 64 neurons. In simulation and on Loihi, the dimensionality was large enough that we could rely upon the pseudo-orthogonality of randomly-generated hypervector representations. This is not reliably achieved for $N=64$, and so we instead manually chose our sparse block code hypervectors $\vec{q}$ and $\vec{b}$ to be orthogonal. The ideal weight matrix $\mat{W}$ was constructed as described in Methods but with the $f$ terms set to $f=0$, to compensate for the fact that our attractor state vectors now have exactly 0 inner product. Since the RRAM devices supported ternary weight states (normalised to $0,0.5,1$), we mapped the ideal weights to ternary weights by applying two thresholds on either side of 0. The distributions of these ternary conductance states are shown in Fig.~\ref{fig:dfa_memristor}.

To store the 64$\times$64 weight matrix on the physical RRAM chip, we partitioned it into two 32$\times$64 matrices and then stored their concatenation on the available 32$\times$128 devices \mbox{(Fig.~\ref{fig:64x64_to_32x128})}. To perform a current readout (whenever a neuron spiked), we input the binary vector of the first 32 neurons as a voltage via the DACs in parallel, and read the first 64 currents. Then, we input the binary vector of the latter 32 neurons' spike activity and added the two current vectors together on the PC. Therefore, all multiply-and-accumulate steps besides the last addition were performed in-memory.

\subsection{Loihi implementation}
The experiments were run on the Intel Loihi 2 Oheo Gulch research system using 1024 neurons (128 blocks of 8 neurons) on 95 neurocores. The code was written in Python using Intel's Lava package.

The network uses the regular current-based leaky integrate and fire neuron model on Loihi.
However, different from the hard-coded model, each neuron has a conventional integrative synapse, with weights set by $\mat{W}$, and a shunting inhibitory synapse to implement the block-WTA and masking behaviour. To achieve the shunting inhibition synapse, we implemented the neuron using custom neuron microcode, a novel feature of Loihi 2.
The microcode on Loihi 2 allows arbitrary instructions from a limited instruction set. To avoid having to create a neuron with 2 registers to store the input, we discriminate the integrative synapse's input and the shunting synapse's input by different weight ranges. The shunting synapse's weight was set to -2048, a value that is never reached by integrating regular input. So, in the microcode, before the regular neuron behaviour is computed, a comparison is made if the input value is below 1000. If it is, the membrane potential is reset to 0 (shunting) and 2048 is added to the input in order not to interfere with the integrative synapse.

The weight matrix was constructed as described in \mbox{Section \ref{sec:W_construction}}, and then quantized by linearly mapping the 4-sigma range of $\mat{W}$ to the available even integer weight values in the interval $[-254, 254]$. These recurrent weights were then written to the all-to-all recurrent synapse weights. Attractor transitions were triggered by stimulating the shunting inhibition synapses, with the input vectors $\vec{s}_0$ or $\vec{s}_1$ to input a 0 or 1, respectively. At the start of the experiment, the RSNN is initialised in the first attractor state $q_0$. Inputs are then given for 200 time steps each, with a 200 time step gap in between, wherein no input is given.

\subsection{Capacity simulation}
% % \color{blue}

The RSNN was modelled as a single-layer fully-connected artificial RNN, with the discrete-time update rule 
\begin{equation}
\vec{z}_{t+1} = \mathrm{bWTA} \Big[ \mat{W} \cdot \big( \vec{z}_t \land \vec{i}_t \big) \Big]
\end{equation}
where $\vec{z}_t \in \{0,1\}^N$ is a sparse block vector representing the network state on time step $t$, $\mat{W}$ is the recurrent weight matrix (see Section~\ref{sec:W_construction}), $\mathrm{bWTA}$ is the block-winner-take-all activation function, and $\vec{i}_t \in \{0, 1\}^N$ is the masking input to the network. If there is no input to the network, then $\vec{i}_t$ is a vector of all ones.

The network was tested with modular-division DFAs of varying sizes $P$, like in Fig.~\ref{fig:sim_walk} for $P = 23$. Note that $P$ does not directly correspond to the number of stored attractor states, since each DFA state actually stores two attractor states and, for these DFAs, two asymmetric transition terms. For various $(P,N)$ pairs, we evaluated the network's ability to correctly emulate the DFA by giving sequences of 5 inputs (each lasting 10 time steps, with 10 time steps of no input between) to the network. A run was deemed to be successful if the hypervector $\vec{q}$ corresponding to the correct final DFA state had the greatest overlap with the network's final state. The block length $L$ was varied such that the active number of neurons $N/L$ scaled with $\log N$, as required to obtain the quadratic capacity relation \cite{knoblauch_memory_2010}. Each $(P,N)$ pair was tested 5 times with different hypervector instantiations.

% \color{black}

\nolinenumbers

\section*{References}
\addcontentsline{toc}{section}{References}

\printbibliography[heading=none]

\end{refsection}

\FloatBarrier

\section*{Acknowledgements}
\addcontentsline{toc}{section}{Acknowledgements}

Thanks to Friedrich Sommer, Denis Kleyko, Chris Kymn and Anthony Thomas for enlightening discussions and suggestions.
This work has been supported by DFG projects NMVAC (432009531) and MemTDE (441959088).
The authors would like to acknowledge the financial support of the CogniGron research center and the Ubbo Emmius Funds (Univ. of Groningen). We thank the Center for Information Technology of the University of Groningen for their support and for providing access to the Hábrók high-performance computing cluster.
We thank Intel for providing access to the Loihi 2 research platform and Philipp Stratmann for support with the microcode.

\section*{Author contributions}

M.C. conceived the initial algorithm and its SNN implementation. H.G., A.R. implemented the algorithm on Loihi. H.W. supervised, and J.C., M.C., implemented the algorithm on the memristive crossbar setup. E.N., H.W., G.I., M.Z., E.C. supervised the experiments. All authors contributed to writing the manuscript.

\section*{Competing interests}
The authors declare no competing interests.

\onecolumn

\newpage

\nolinenumbers

\begin{center}
\textsc{Supplementary Material}
\end{center}

\begin{refsection}

\renewcommand\thefigure{S\arabic{figure}}
\renewcommand\thesection{S\arabic{section}}
\renewcommand{\theHfigure}{S\arabic{figure}}
\renewcommand\theHsection{S\arabic{section}}

\setcounter{figure}{0}
\setcounter{section}{0}
\setcounter{equation}{0}
\setcounter{page}{1}

\section{Sparse block code overlaps}
\label{sec:sbc_capacity}

We here derive a result, adapted from~\cite{thomas_theoretical_2022}, that the number of sparse block code (SBC) hypervectors that can be randomly generated without accidentally incurring a significant degree of overlap between any pair of them, is exponential in dimension $N$, and so very large for reasonable values of $N$ and $L$.

We generate SBC hypervectors of dimension $N$, with block length $L$, and number of blocks $M = N/L$. The coding level is then $f = 1/L$. The normalised similarity between SBC hypervectors is given by
\begin{equation}
    \mathrm{sim}(\vec{a}, \vec{b}) = \frac{1}{M} \vec{a} \cdot \vec{b}
\end{equation}
such that the the similarity between identical hypervectors $\mathrm{sim}(\vec{a}, \vec{a}) = 1$, and the expected similarity between independent hypervectors is $\langle \mathrm{sim}(\vec{a}, \vec{b}) \rangle = f < 1$.

We first provide a bound on the similarity between independent hypervectors. We can rewrite the similarity between SBC hypervectors as
\begin{equation}
    \mathrm{sim}(\vec{a}, \vec{b}) = \frac{1}{M} \sum_{i=1}^{M} X_i
\end{equation}
where $X_i \in \{0,1\}$ denotes whether the same neuron in the $i$'th block is active in both hypervectors, with $\langle X_i \rangle = f$. By Hoeffding's inequality, this sum is bounded by
\begin{equation}
    \mathbb{P}\big[ \mathrm{sim}(\vec{a}, \vec{b}) - f \geq \theta \big] \leq  \exp \big( -2 M \theta^2 \big)
\end{equation}
where $\theta$ is a threshold of our choosing. This bound gives us assurances about how much the similarity between independent hypervectors can deviate from the expected value $f$. We can then set $\theta$ to represent the maximum acceptable similarity between independent hypervectors. Given that independent hypervectors have expected similarity $f$, and identical hypervectors have similarity $1$, then a reasonable threshold might be halfway between them, $\theta = \frac{1}{2}(1-f)$. For reasonable values of $N = 1000$, $L = 8$, the probability that a given pair of independent hypervectors exceeds $\theta$ is then less than $10^{-20}$.

We however want a bound on the number $K$ of SBC hypervectors that can be generated, for which \textit{none} of the possible inner product pairs exceed $\theta$. For $K$ hypervectors, there are $\frac{1}{2}K(K-1) < \frac{1}{2} K^2$ pairs to check. We apply Boole's inequality, that the union probability of a set of events occurring $\mathbb{P} \big[ \bigcup_i E_i \big] \leq \sum_i \mathbb{P}[E_i]$ for some countable events $\{E_i\}$. In our case, each event is whether a similarity exceeds $\theta$, and so we have
\begin{equation}
\mathbb{P}[\text{any pair exceeds $\theta$}] \leq \delta  = \sum_{j = 1}^{\frac{1}{2}K^2} \exp \big( -2 M\theta^2 \big) = \frac{1}{2}K^2\exp \big( -2 M\theta^2 \big)
\end{equation}
where we have introduced $\delta \in [0,1]$ as the acceptable probability of a collision, which we will set to $\delta \ll 1$. We can then rearrange for the number of hypervectors $K$,
\begin{equation}
    K = \sqrt{2 \delta} \exp \Big( M\theta^2 \Big) = \sqrt{2 \delta} \exp \Big( N\theta^2 /L \Big)
\end{equation}
and so the number of SBC hypervectors that can be generated before a collision becomes likely, is exponential in dimension $N$. For comparison, the number of orthogonal localist representations that can be generated is linear in $N$.

For reasonable values of $N = 1000$, $L=8$, with an acceptable failure probability of $\delta = 0.0001$, then with probability greater than $0.9999$, we can generate $K \approx 3 \times 10^9$ SBC hypervectors before a collision occurs. This is well in excess of the number of states that can be stored in a fully-connected attractor network, which scales at most with the number of synapses $N^2$.

\section{Postsynaptic sum calculations}

\label{sec:h_calc}
\FloatBarrier

We first calculate the mean and variance of the components of the postsynaptic sum $\vec{h}$, in the case that the network is already in state $\vec{q}^\mu \in Q$, and there is no input. We calculate the contribution to $\vec{h}$ due to the three superimposed matrices in $\mat{W}$ separately, for the sake of cleanliness. We will assume that the inner products between different hypervectors are independent (while they are actually only pairwise independent), such that their variances combine linearly.
\begin{equation}
\begin{split}
    [\mat{W}_\text{attr.} \vec{q}^\mu]_i & = \sum_\nu^Q (q_i^\nu -f)(\vec{q}^\nu -f ) \cdot \vec{q}^\mu \\
    & = M(1-f)(q_i^\mu -f) + \sum_{\nu \neq \mu}^Q \underbrace{(q_i^\nu -f)}_{\substack{\text{Mean = 0} \\ \text{Var. = } f(1-f)}}\underbrace{(\vec{q}^\nu -f ) \cdot \vec{q}^\mu}_{\substack{\text{Mean = 0} \\ \text{Var. = } Mf(1-f)}} \\
    & =  M(1-f)(q_i^\mu -f) \pm f(1-f)\sqrt{(Q-1)M} \\
    & \approx  M(q_i^\mu -f) \pm f\sqrt{MQ}
\end{split}
\end{equation}
where we are abusing notation somewhat by using $Q$ to denote the number of states in the set $Q$ (and will do the same with $E$ and $S$), and on the last line we make the approximation that $Q \gg 1$ (there are many many DFA states) and $f \ll 1$ (our states are sparse). The corresponding calculation for the bridge matrix is
\begin{equation}
\begin{split}
    [\mat{W}_\text{brdg.} \vec{q}^\mu]_i & = \sum_{\nu}^Q \Big[ \underbrace{(q_i^\nu -f)}_{\substack{\text{Mean = 0} \\ \text{Var. = } f(1-f)}}\underbrace{(\vec{b}^\nu -f ) \cdot \vec{q}^\mu}_{\substack{\text{Mean = 0} \\ \text{Var. = } Mf(1-f)}} + \sum_s^S \underbrace{(b_i^\nu -q_i^\nu)}_{\substack{\text{Mean = 0} \\ \text{Var. = } 2f(1-f)}}\underbrace{((\vec{b}^\nu -f) \had \overline{\vec{s}} ) \cdot \vec{q}^\mu}_{\substack{\text{Mean = 0} \\ \text{Var. = } Mf(1-f)}} \Big] \\
    & = 0 \pm f(1-f) \sqrt{MQ(1+2S)} \\
    & \approx 0 \pm f \sqrt{MQ(1+2S)} \\
\end{split}
\end{equation}
If we are constructing a DFA with many inputs (large $S$), but each state has incoming edges with only a few of these inputs, then the capacity of the network can be increased by here summing not over all inputs, but only over the unique inputs of the edges to the state $q$. The transition matrix calculation goes
\begin{equation}
\begin{split}
    [\mat{W}_\text{tran.} \vec{q}^\mu]_i & = \sum_\eta^E (b_i^{\prime} -q_i ) ((\vec{q} -f) \had \overline{\vec{s}} ) \cdot \vec{q}^\mu \\
    & = \sum_{\eta | q \neq q^\mu}^E \underbrace{(b_i^{\prime} -q_i )}_{\substack{\text{Mean = 0} \\ \text{Var. = } 2f(1-f)}}\underbrace{((\vec{q} -f) \had \overline{\vec{s}} ) \cdot \vec{q}^\mu}_{\substack{\text{Mean = 0} \\ \text{Var. = } Mf(1-f) }} + \sum_{\eta | q = q^\mu}^E \underbrace{(b_i^{\prime} -q_i)}_{\substack{\text{Mean = 0} \\ \text{Var. = } 2f(1-f)}}\underbrace{((\vec{q}^\mu -f) \had \overline{\vec{s}} ) \cdot \vec{q}^\mu}_{\substack{\text{Mean = 0} \\ \text{Var. = } M(1-f)^2 }} \\
    & = 0 \pm (1-f)\sqrt{2Mf^2(E-S) + 2MSf(1-f)} \\
    & \approx 0 \pm \sqrt{2Mf^2E + 2MSf} \\
\end{split}
\end{equation}
where we split the summation up into transitions whose source state is $q^\mu$ (and those with a different source state). We then assume the worse-case scenario that $S$ of the $E$ edges have $q^\mu$ as a source state. The final postsynaptic sum is given by
\begin{equation}
\begin{split}
[\vec{h}]_i = [\mat{W} \vec{q}^\mu]_i & \approx M(q_i^\mu -f) \pm \sqrt{M}\sqrt{2f^2Q(1+S)+ 2f^2E + 2fS} \\
% & \propto q_i^\mu -f \pm \frac{1}{\sqrt{M}} \sqrt{f^2Q + 2f^2QS + 2f^2E + 2Sf} \\
& \propto q_i^\mu -f \pm \sqrt{\frac{2}{N}} \sqrt{fQ(1+S)+ fE + S} \\
\end{split}
\end{equation}
from which we get $\langle \mat{W}\vec{q}^\mu \rangle \propto \vec{q}^\mu -f$.

We then calculate the postsynaptic sum when we are masking the network with an input $s^\lambda$, and we assume there exists an edge $\epsilon \in E$ with source state $q^\mu$ and input $s^\lambda$. Calculating $\vec{h}$ in the same way, we have
\begin{equation}
\begin{split}
[\mat{W}_\text{attr.} (\vec{q}^\mu \land \vec{s}^\lambda)]_i & = \sum_\nu^Q (q_i^\nu -f)(\vec{q}^\nu -f ) \cdot (\vec{q}^\mu \land \vec{s}^\lambda) \\
    & = \frac{1}{2}M(1-f)(q_i^\mu -f) + \sum_{\nu \neq \mu}^Q \underbrace{(q_i^\nu -f)}_{\substack{\text{Mean = 0} \\ \text{Var. = } f(1-f)}}\underbrace{(\vec{q}^\nu -f ) \cdot (\vec{q}^\mu \land \vec{s}^\lambda)}_{\substack{\text{Mean = 0} \\ \text{Var. = } \frac{1}{2}Mf(1-f)}} \\
    & =  \frac{1}{2} M(1-f)(q_i^\mu -f) \pm f(1-f)\sqrt{(Q-1)\frac{1}{2}M} \\
    & \approx  \frac{1}{2}M(q_i^\mu -f) \pm f\sqrt{\frac{1}{2}MQ}
\end{split}
\end{equation}
and for the bridge matrix
\begin{equation}
\begin{split}
    [\mat{W}_\text{brdg.} (\vec{q}^\mu \land \vec{s}^\lambda)]_i & = \sum_{\nu}^Q \Big[ \underbrace{(q_i^\nu -f)}_{\substack{\text{Mean = 0} \\ \text{Var. = } f(1-f)}}\underbrace{(\vec{b}^\nu -f ) \cdot (\vec{q}^\mu \land \vec{s}^\lambda)}_{\substack{\text{Mean = 0} \\ \text{Var. = } \frac{1}{2}Mf(1-f)}} + \sum_s^S \underbrace{(b_i^\nu -q_i^\nu)}_{\substack{\text{Mean = 0} \\ \text{Var. = } 2f(1-f)}}\underbrace{((\vec{b}^\nu -f) \had \overline{\vec{s}} ) \cdot (\vec{q}^\mu \land \vec{s}^\lambda)}_{\substack{\text{Mean = 0} \\ \text{Var. = } \frac{1}{2}Mf(1-f)}} \Big] \\
    & = 0 \pm f(1-f) \sqrt{\frac{1}{2}MQ(1+2S)} \\
    & \approx 0 \pm f \sqrt{\frac{1}{2}MQ(1+2S)} \\
\end{split}
\end{equation}
and the transition matrix
\begin{equation}
\begin{split}
& [\mat{W}_\text{tran.}   (\vec{q}^\mu \land \vec{s}^\lambda)]_i  = \sum_\eta^E (b_i^\prime - q_i) ((\vec{q} - f) \had \overline{\vec{s}} ) \cdot (\vec{q}^\mu \land \vec{s}^\lambda) \\
& = (1-f)\frac{M}{2}(b_i^\prime - q_i^\mu) + \sum_{\substack{\eta\neq \epsilon \\ \text{and } q \neq q^\mu}}^E \underbrace{(b_i^\prime - q_i)}_{\substack{\text{Mean = 0} \\ \text{Var. = } 2f(1-f)}} \underbrace{((\vec{q} - f) \had \overline{\vec{s}} ) \cdot (\vec{q}^\mu \land \vec{s}^\lambda)}_{\substack{\text{Mean = 0} \\ \text{Var. = } \frac{M}{2} f(1-f) }} + \sum_{\substack{\eta\neq \epsilon \\ \text{and } q = q^\mu}}^E \underbrace{(b_i^\prime - q_i)}_{\substack{\text{Mean = 0} \\ \text{Var. = } 2f(1-f)}} \underbrace{((\vec{q} - f) \had \overline{\vec{s}} ) \cdot (\vec{q}^\mu \land \vec{s}^\lambda)}_{\substack{\text{Mean = 0} \\ \text{Var. = } \frac{M}{2} (1-f)^2 }}  \\
& = \frac{1}{2} M (1-f) (b_i^\prime - q_i^\mu) \pm (1-f)\sqrt{ M  \big( f^2 (E-S+1) + (S-1)f(1-f) \big)} \\
& \approx \frac{1}{2} M (b_i^\prime - q_i^\mu) \pm \sqrt{M( f^2E + f(S-1))}
\end{split}
\end{equation}
where we split the summation over all transitions $E$ into the correct transition $\epsilon$, incorrect transitions with different source states, and incorrect transitions with the same source state. We then assume the worst case where there are $S-1$ incorrect transitions with the same source state.

Combining the three again, we get
\begin{equation}
\begin{split}
[\vec{h}]_i = [\mat{W} (\vec{q}^\mu \land \vec{s}^\lambda)]_i & \approx \frac{M}{2} (b_i^\prime - f) \pm \sqrt{M}\sqrt{f^2 Q(1+S) + f^2E + f(S-1)} \\
% & \propto (b_i^\prime -f ) \pm \frac{2}{\sqrt{M}} \sqrt{f^2 Q(1+S) + f^2E + f(S-1)}
& \propto (b_i^\prime -f ) \pm \frac{2}{\sqrt{N}} \sqrt{f Q(1+S) + fE + S-1}
\end{split}
\end{equation}
which in expectation gives us $\langle \mat{W} (\vec{q}^\mu \land \vec{s}^\lambda) \rangle \propto \vec{b}^\prime -f$. The same procedure can be applied to calculate the mean and variance of the postsynaptic sum in all cases.

\section{Energy function}

\label{sec:energy}

We consider a simplification of the network dynamics, where each block is updated asynchronously in discrete time steps, in which one block is picked at random, and the neuron with the greatest input spikes.
The RSNN can then be coarsely described by a neural state vector $\vec{z}_t \in \{0, 1\}^N$, indicating which neuron is active in each block on time step $t$. If we are in a regime where we can ignore the effects of the bridge and transition matrices, then we can make the substitution $\mat{W} = \mat{W}^\text{attr.}$. We can split the neural state vector $\vec{z}_t$ into the vector describing only the components in the updating block, $\vec{z}^b_t$, and the vector describing all the other $N-L$ components $\vec{z}^{\neg b}_t$, with $\vec{z}_t = \vec{z}^b_t + \vec{z}^{\neg b}_t$. If we define the energy $E_t$ at time step $t$ as $E_t = - \vec{z}_t\T \mat{W} \vec{z}_t$, then the change in energy $\Delta E_{t+1}  = E_{t+1} - E_t$ is given by
\begin{equation}
    \Delta E_{t+1} = -(\vec{z}_{t+1}^{b} - \vec{z}_{t}^{b})\T \mat{W} \vec{z}_t^{\neg b}
\end{equation}
where we have used that $\mat{W}$ is symmetric, and there are no interactions (weight is 0) between neurons in the same block~\cite{cohen_absolute_1983}. Now, the block-WTA dynamics mean that each block effectively computes an $\mathrm{argmax}$ over its inputs, meaning that $\vec{z}_{t+1}^{b}$ maximises the inner product $\vec{z}_{t+1}^{b \intercal} \mat{W} \vec{z}_t^{\neg b}$, and so is the maximum possible value that $\vec{z}_{t}^{b \intercal} \mat{W} \vec{z}_t^{\neg b}$ can take. Thus, we have $\Delta E_t \leq 0 $, i.e. the network dynamics descend the energy surface $E(\vec{z})$, which by construction
\begin{equation}
E(\vec{z}) = -\vec{z}\T \mat{W}_\text{attr.} \vec{z}  = -\sum_{q \in Q} \big( \vec{z} \cdot (\vec{q} -f) \big)^2
\end{equation}
consists of parabolic energy wells at our DFA states $\vec{q}$.

\section{Mixing SBC and MAP VSA models}
\label{sec:sm_sbc_map}

% \color{blue}

\begin{figure}
\centering
\includegraphics[width=5.2in]{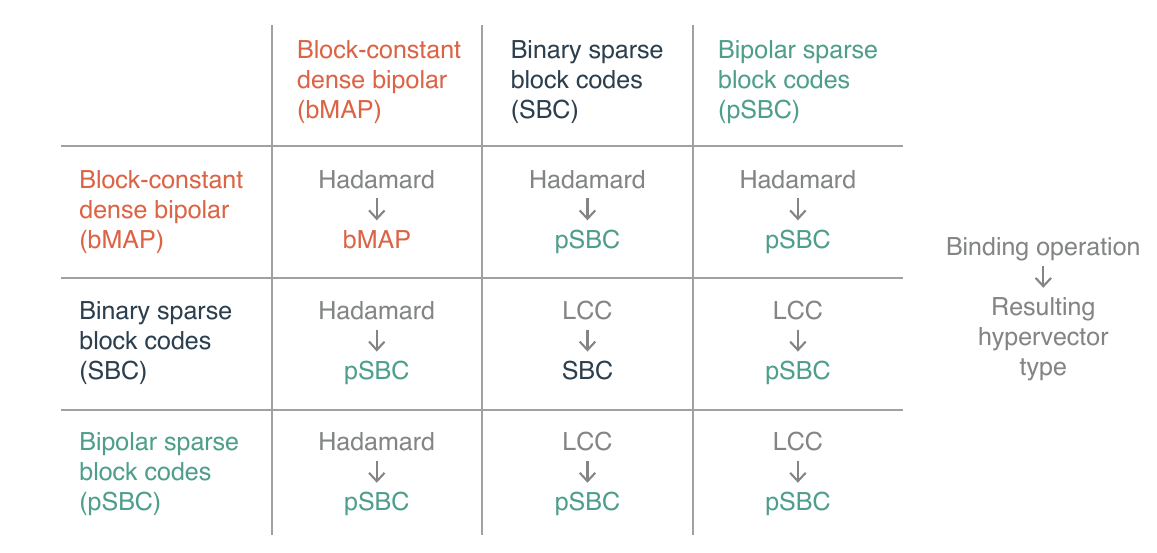}
\caption{Suitable binding operations between hypervectors of different type, as well as the type of hypervector it produces.}
\label{fig:sm_vsa_table}
\end{figure}

In this work, we made simultaneous use of both the binary Sparse Block Code (SBC) \cite{laiho_high-dimensional_2015, frady_variable_2023} and Multiply-Add-Permute (MAP) \cite{gayler_multiplicative_1998, kleyko_survey_2022} VSA models, with the latter model using dense bipolar hypervectors. Contrary to VSA orthodoxy, we utilised the MAP binding operation (Hadamard product, denoted ``$\had$'') between both hypervector types in the construction of the recurrent weight matrices. We did not make use of the conventional binding operation with SBC hypervectors, blockwise Local Circular Convolution (LCC), denoted ``$\LCC$''. Here, we briefly explore the system of hypervectors that arises when mixing the MAP and SBC models.

We used the Hadamard binding operation to bind dense bipolar MAP hypervectors to SBC hypervectors, as this produced \textit{bipolar} SBC hypervectors which were pseudo-orthogonal to the binary SBC hypervectors representing the network state at any time. The dense bipolar MAP hypervectors used here needed to be block-constant: only $N/L$ random samples of $\{-1, +1\}$ were drawn per hypervector, one for each block, and all components in the same block took this value.

In our experiments, this was required such that an effective unbinding of the MAP vectors could be performed by masking out entire blocks. Masking out only some neurons in each block (if the MAP vectors were not block-constant) conflicts with the winner-take-all mechanism led to the incorrect states being represented.
For the joint MAP-SBC hypervector system, the MAP hypervectors must be block-constant to ensure the Hadamard and LCC binding operations are associative. Otherwise, with unconstrained MAP hypervectors, one would have
\begin{equation}
     \big( \vec{MAP} \had \vec{SBC}_1 \big) \LCC  \vec{SBC}_2 \neq  \vec{MAP} \had \big( \vec{SBC}_1 \LCC  \vec{SBC}_2 \big)
\end{equation}
which is not a desirable property with current VSA approaches, not least of all because unbinding would then require information about the previous order of binding operations. For clarity, and to differentiate unconstrained dense bipolar hypervectors from block-constant dense bipolar hypervectors, we refer to the latter as \textit{bMAP} hypervectors.

Hadamard binding a bMAP hypervector with an SBC hypervector produces a bipolar sparse block code hypervector, which we refer to as \textit{pSBC} hypervectors, to differentiate them from the binary SBC hypervectors. The binding here must be Hadamard rather than LCC since LCC binding would produce a dense hypervector with all indices in each block having the same value, and so the SBC hypervector would not be at all recoverable. The pSBC hypervector produced by Hadamard binding bMAP and SBC hypervectors is then pseudo-orthogonal to the SBC vector, but not to the bMAP vector, the consequences of which we will shortly discuss. Independently-generated pSBC hypervectors are then pseudo-orthogonal to each-other (they have 0 expected inner product), can be bound together by LCC binding, and unbound using inverse-LCC binding. A summary of the suitable binding operations between the three hypervector types (bMAP, SBC, pSBC) is shown in Fig.~\ref{fig:sm_vsa_table}.

pSBC hypervectors can thus be used in the same way as SBC hypervectors, with binding by LCC and superposition by element-wise addition, but with the added ease that independently-generated pSBC hypervectors have zero expected inner product.

The more interesting case is not to consider pSBC hypervectors that are independently generated, but rather which exist as a by-product of the simultaneous use of bMAP and SBC hypervectors. Since the bMAP hypervectors are constant-valued in each block, under LCC binding their effect amounts simply to a prefactor in each block, and so Hadmard binding commutes with LCC binding, i.e.
\begin{equation}
\begin{split}
\vec{pSBC}_1 \LCC \vec{pSBC}_2 & = \big( \vec{bMAP}_1 \had \vec{SBC}_1 \big) \LCC \big( \vec{bMAP}_2 \had \vec{SBC}_2 \big) \\
& = \big( \vec{bMAP}_1 \had \vec{bMAP}_2 \big) \had \big( \vec{SBC}_1 \LCC \vec{SBC}_2 \big) \\
& = \vec{bMAP}_3 \had \vec{SBC}_3 \\
& = \vec{pSBC}_3
\end{split}
\end{equation}

Hence, LCC binding with pSBC hypervectors can be seen as actually performing two separate binding operations in the bMAP and SBC domains.

To exemplify the use of pSBC hypervectors and to empirically demonstrate viability as a VSA framework, we turn to the commonly used VSA analogical reasoning tutorial ``What is the dollar of Mexico?'' \cite{kanerva_hyperdimensional_2009}. There, two hypervectors $\vec{usa}$ and $\vec{mex}$ represent information about the USA and Mexico respectively, by superimposing bound role-filler pairs created from the atomic hypervectors $\vec{cap}, \vec{cur}, \vec{wdc}, \vec{dol}, \vec{mxc}$ and $\vec{pes}$ representing ``capital city'', ``currency'', ``Washington DC'', ``dollar'', ``Mexico City'' and ``peso'' respectively \cite{kanerva_hyperdimensional_2009}.

\subsection{Case 1: All pSBC}
We consider three cases. First, that all the atomic hypervectors are independently-generated pSBC hypervectors. The USA hypervector is then formed by LCC binding the corresponding role-filler hypervector pairs,
\begin{equation}
    \vec{usa}_\text{pSBC} = \vec{cap}_\text{pSBC} \LCC \vec{wdc}_\text{pSBC} +  \vec{cur}_\text{pSBC} \LCC \vec{dol}_\text{pSBC}
\end{equation}
from which one can decode the currency used in the USA by unbinding the currency hypervector via inverse LCC: The only hypervector with significant overlap with $\vec{usa}_\text{pSBC} \LCC^{-1} \vec{cur}_\text{pSBC}$ is $\vec{dol}_\text{pSBC}$, the correct answer (Fig.~\ref{fig:sm_mexico}a). If $\vec{mex}_\text{pSBC}$ is constructed analagously to $\vec{usa}_\text{pSBC}$, the ``What's the dollar of Mexico'' example can be performed by calculating $\vec{mex}_\text{pSBC} \LCC^{-1} \vec{usa}_\text{pSBC} \LCC \vec{dol}_\text{pSBC}$, which is similar only to $\vec{pes}_\text{pSBC}$, as desired (Fig.~\ref{fig:sm_mexico}d). 

\subsection{Case 2: Roles SBC, fillers bMAP}

In the second case, we consider that the roles (currency, capital) are represented by SBC hypervectors, and the fillers (Washington DC, dollar, $\ldots$) are represented by bMAP hypervectors. The USA hypervector is then constructed as
\begin{equation}
\vec{usa}_\text{pSBC} = \vec{cap}_\text{SBC} \had \vec{wdc}_\text{bMAP} +  \vec{cur}_\text{SBC} \had \vec{dol}_\text{bMAP}
\end{equation}
Decoding the currency of the USA is performed by unbinding the currency hypervector, again by inverse LCC. The resulting expression reads
\begin{equation}
\begin{split}
\vec{usa}_\text{pSBC} \LCC^{-1} \vec{cur}_\text{SBC} & = \big( \vec{cap}_\text{SBC}\LCC^{-1} \vec{cur}_\text{SBC} \big) \had \vec{wdc}_\text{bMAP} + \big( \vec{cur}_\text{SBC} \LCC^{-1} \vec{cur}_\text{SBC} \big) \had \vec{dol}_\text{bMAP} \\
& = \big( \vec{cap}_\text{SBC}\LCC^{-1} \vec{cur}_\text{SBC} \big) \had \vec{wdc}_\text{bMAP} + \vec{1}_\text{SBC} \had \vec{dol}_\text{bMAP}
\end{split}
\end{equation}
where $\vec{1}_\text{SBC}$ is the identity hypervector under LCC, given by a hypervector where only the first element in each block is active. Hence, to correctly calculate similarities with the bMAP hypervectors, they should first be bound to the SBC identity vector. Results are shown in Fig.~\ref{fig:sm_mexico}b. Note that, although the $\vec{dol}_\text{bMAP}$ hypervector correctly has the greatest similarity, the $\vec{wdc}_\text{bMAP}$ hypervector also has a notable similarity, despite being bound to two SBC hypervectors. This is because the SBC hypervectors are not balanced, consisting only of non-negative values, and so the similarity with the $\vec{wdc}_\text{bMAP}$ is reduced by a factor of $1/L$, rather than to 0. A similar effect would be observed if another bound copy of $\vec{dol}_\text{bMAP}$ were superimposed, but bound to a different SBC hypervector.

Thus, unlike most VSA approaches, this binding method has produced a hypervector which retains similarity with one of its bound hypervectors, while still unambiguously coding for the binding of the two symbols. Although this raises the possibility of false positives, it opens an interesting avenue for future research exploring whether such properties could be used beneficially.

\subsection{Case 3: Roles bMAP, fillers SBC}

\begin{figure}
\centering
\includegraphics[width=4.5in]{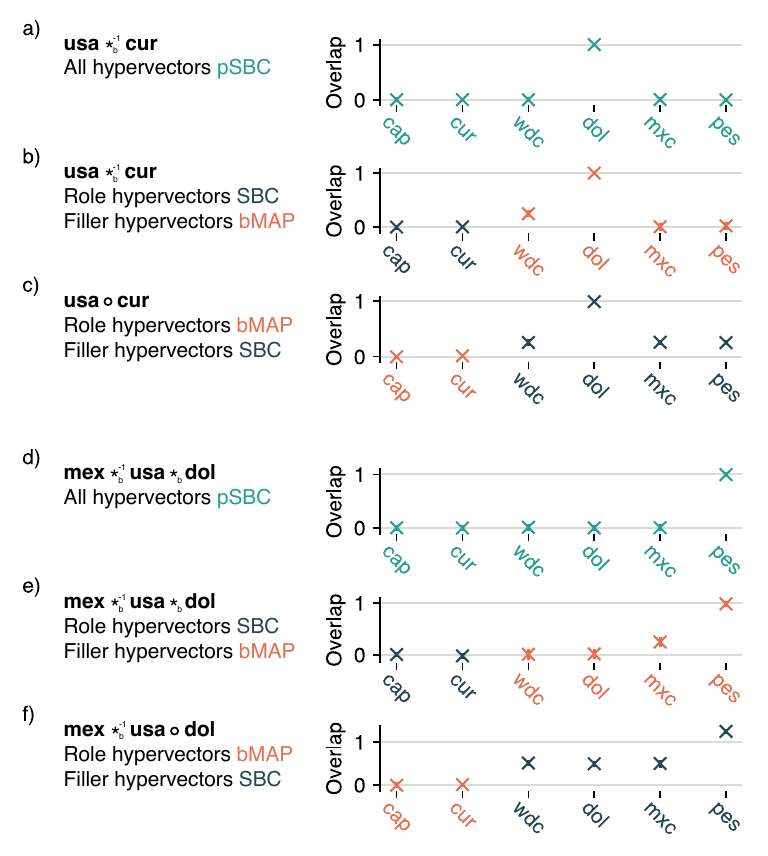}
\caption{Hypervector decoding with different combinations of SBC, bMAP, and pSBC hypervectors. \textbf{a)} Inner products between the atomic hypervectors and the hypervector formed by unbinding the currency hypervector from the USA hypervectors, if all atomic hypervectors are independently-generated pSBC hypervectors, or \textbf{b)} the role and filler hypervectors are SBC and bMAP hypervectors respectively, or \textbf{c)} vice versa. \textbf{d-f}) The same choices of hypervectors, put performing the ``What's the dollar of Mexico?'' decoding task \cite{kanerva_hyperdimensional_2009}. Values used $N = 1024$, $L = 4$, 100 trials.}
\label{fig:sm_mexico}
\end{figure}

In the third case, the roles are represented by bMAP hypervectors and the fillers by SBC hypervectors. To decode the currency of the USA, we unbind the currency hypervector $\vec{cur}_\text{bMAP}$, now via a Hadamard product, resulting in
\begin{equation}
\vec{usa}_\text{pSBC} \had \vec{cur}_\text{bMAP} = \vec{cap}_\text{bMAP} \had \vec{cur}_\text{bMAP} \had \vec{wdc}_\text{SBC} + \vec{dol}_\text{SBC}
\end{equation}
This expression is maximally similar to the $\vec{dol}_\text{SBC}$ as desired (Fig.~\ref{fig:sm_mexico}c). However, due to the presence of an unbalanced SBC hypervector, this expression will also have a similarity of approximately $1/L$ with all the other SBC vectors.

Computing the full ``What's the dollar of Mexico?'' analogy as in \cite{kanerva_hyperdimensional_2009} requires binding the USA to Mexico mapping to dollar $ \vec{mex}_\text{pSBC} \LCC^{-1} \vec{usa}_\text{pSBC} \LCC \vec{dol}_\text{SBC}$, which reads:
\begin{equation}
\vec{mxc}_\text{SBC} \LCC^{-1} \vec{wdc}_\text{SBC} \LCC \vec{dol}_\text{SBC} +  \vec{cap}_\text{bMAP} \had \vec{cur}_\text{bMAP} \big( \vec{mxc}_\text{SBC} + \vec{pes}_\text{SBC} \LCC^{-1} \vec{wdc}_\text{SBC} \LCC \vec{dol}_\text{SBC} \big) + \vec{pes}_\text{SBC}
\end{equation}

The first set of terms forms an SBC hypervector, that will have similarity $1/L$ with any of our SBC hypervectors. The second set of terms is a balanced pSBC hypervector, pseudo-orthogonal to all our atomic hypervectors. Since the only unbound term is the peso, this expression is maximally similar to $\vec{pes}_\text{SBC}$, as desired. However, it will have an overlap of $1 + 1/L$ due to the expected overlap with the first set of terms (Fig.~\ref{fig:sm_mexico}f).

\subsection{Should multiple VSA models be explored?}

When making simultaneous use of multiple VSA models, certain considerations must thus be taken into account. First, different hypervector types may affect what similarity structure is exposed/hidden through binding. Second, prior knowledge about the structure of a hypervector may be required to ascertain which binding operations are suitable. These requirements arguably detract from the self-consistency and simplistic elegance of VSAs that make them so attractive as a unified system of representation.

On the other hand, if one believes that the brain may utilise the principles of VSAs, it is nonetheless unlikely that a single common choice of representation would be used throughout. For this reason alone, it is worth considering how VSAs perform when multiple representation schemes are at play, and what additional algebraic structure is introduced.
It should be explored further what general properties arise when multiple VSA models are integrated together, potentially operating on different parts of the hypervector or at different scales, and how this additional representational flexibility can be effectively utilised.

% \color{black}

\FloatBarrier

\begin{figure}
    \centering
    \includegraphics[width=\linewidth]{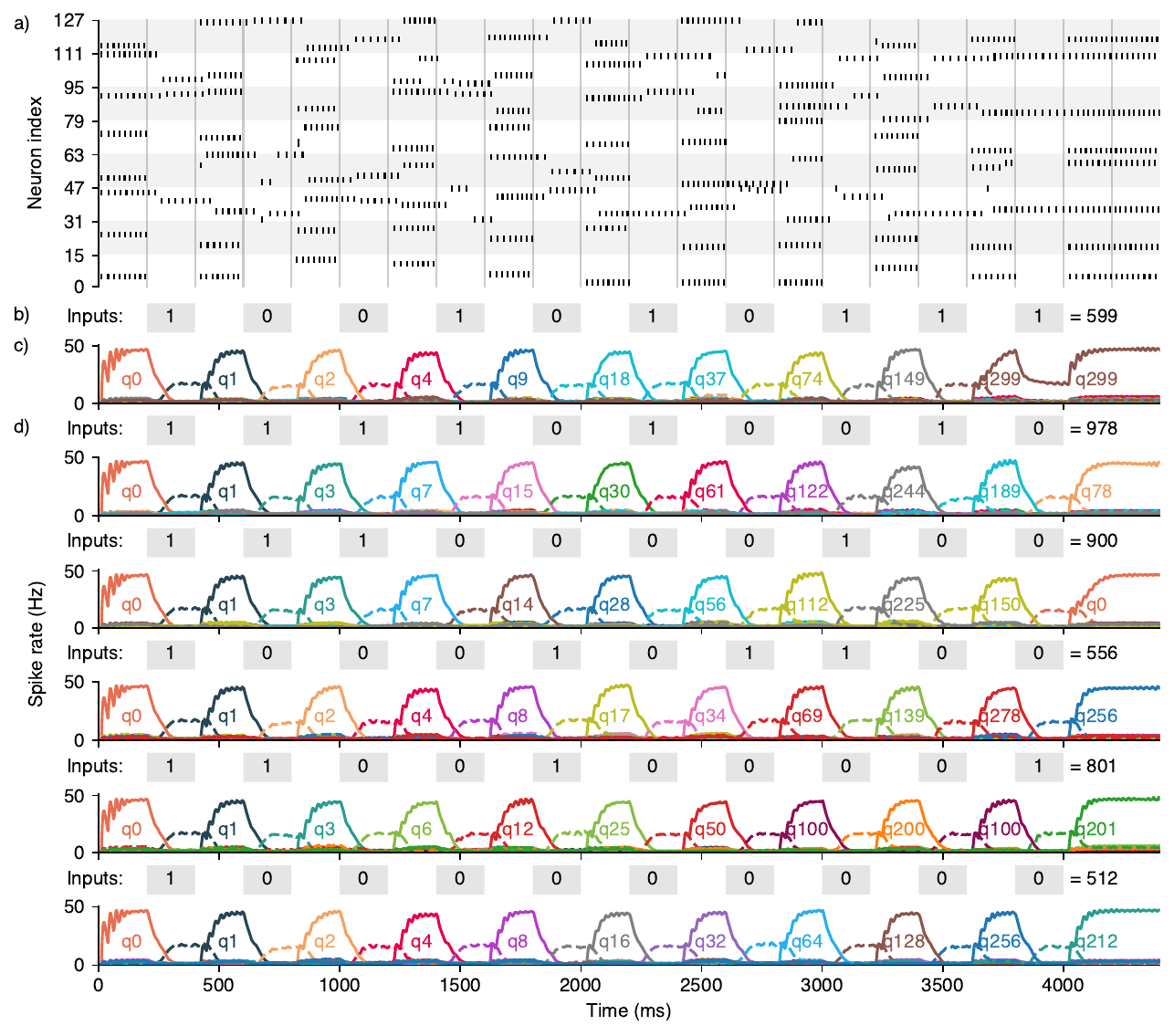}
    \caption{A 300-state DFA embedded into an RSNN in simulation with ideal weights. The chosen DFA implements modular division of binary numbers by 300, generated in the same way as in Fig.~\ref{fig:sim_walk}. \textbf{a)} A raster plot of the first 64 neurons' spikes, with block length $L = 16$. \textbf{b)} The symbolic inputs to the RSNN. \textbf{c)} The mean firing rate of the neurons in each attractor state. \textbf{d)} Different input sequences are given to otherwise identical simulations. In all cases, the RSNN performs the correct walk between attractor states, with the final inhabited state indicating the result of the input binary number $\mathrm{mod} \ 300$. By specifying our RSNN dynamics in the VSA framework, scaling up the algorithm to large state machines is seamless, and is limited only by the memory capacity of the attractor network.}
    \label{fig:sim_huge_dfa_results}
\end{figure}

\begin{figure}
    \centering
    \includegraphics[width=0.9\linewidth]{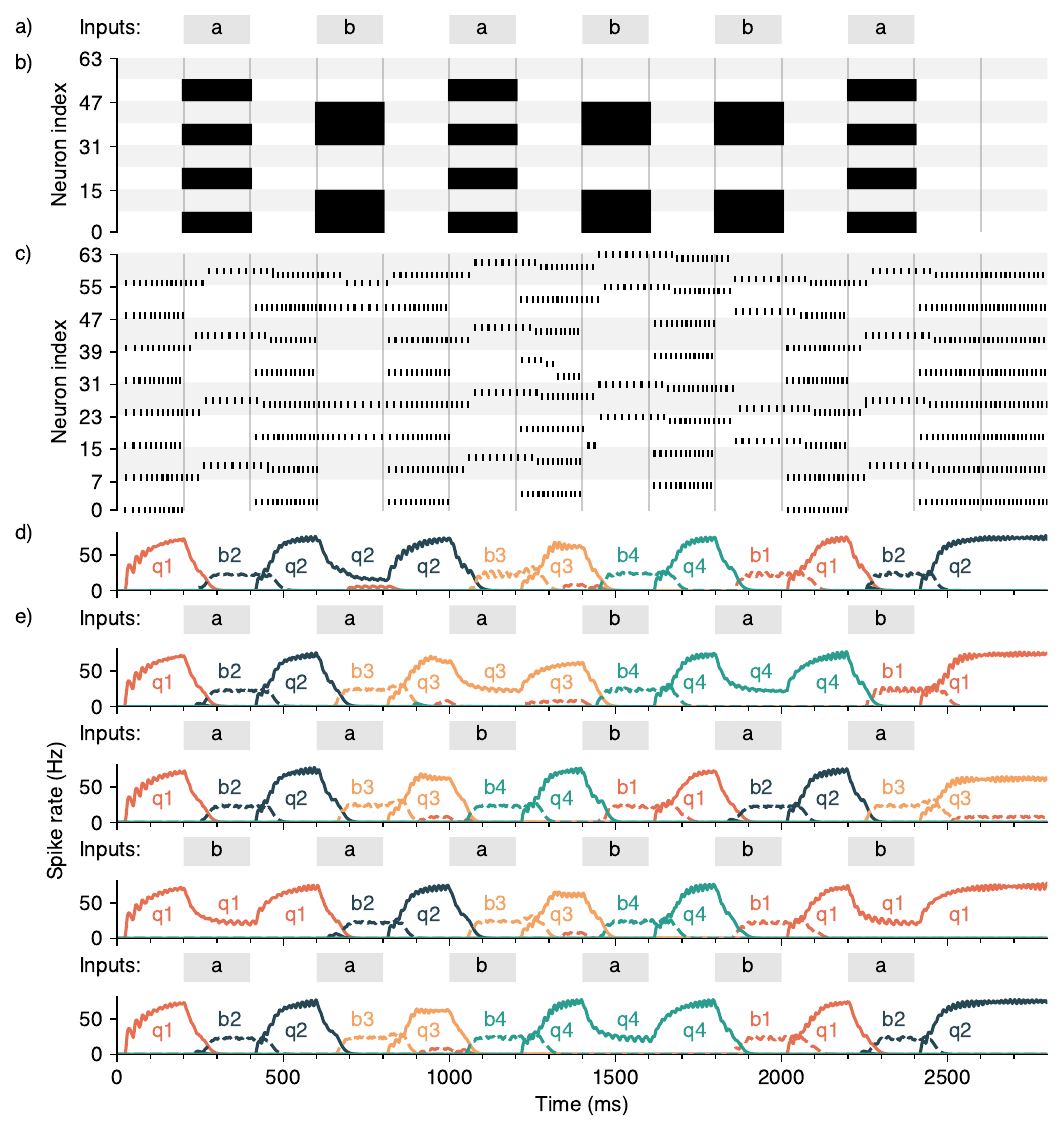}
    \caption{The closed-loop memristive crossbar experiment, configured to embed a second DFA. The DFA is given by $q_0 \overset{a}{\rightarrow} q_1 \overset{a}{\rightarrow} q_2 \overset{b}{\rightarrow} q_3 \overset{b}{\rightarrow} q_0$. \textbf{a)} The symbolic input to the network. \textbf{b)} The masking input to the network, with a separate mask for $a$ and $b$. Due to size constraints, the mask vectors were chosen to be orthogonal.  \textbf{c)} The spiking activity of neurons within the RSNN. \textbf{d)} The mean firing rates of the neurons within each attractor state. \textbf{e)} The same experiment was performed with different sequences of inputs, which cause corresponding different walks between the states. The RSNN performs the correct walk between attractor states in all cases. Although there is erroneous neuron firing in some blocks (due to the nonideal device conductances), the distributed states ensure the overall dynamics are unaffected.}
    \label{fig:xbar_bigger}
\end{figure}

\begin{figure}
    \centering
    \includegraphics[width=5in]{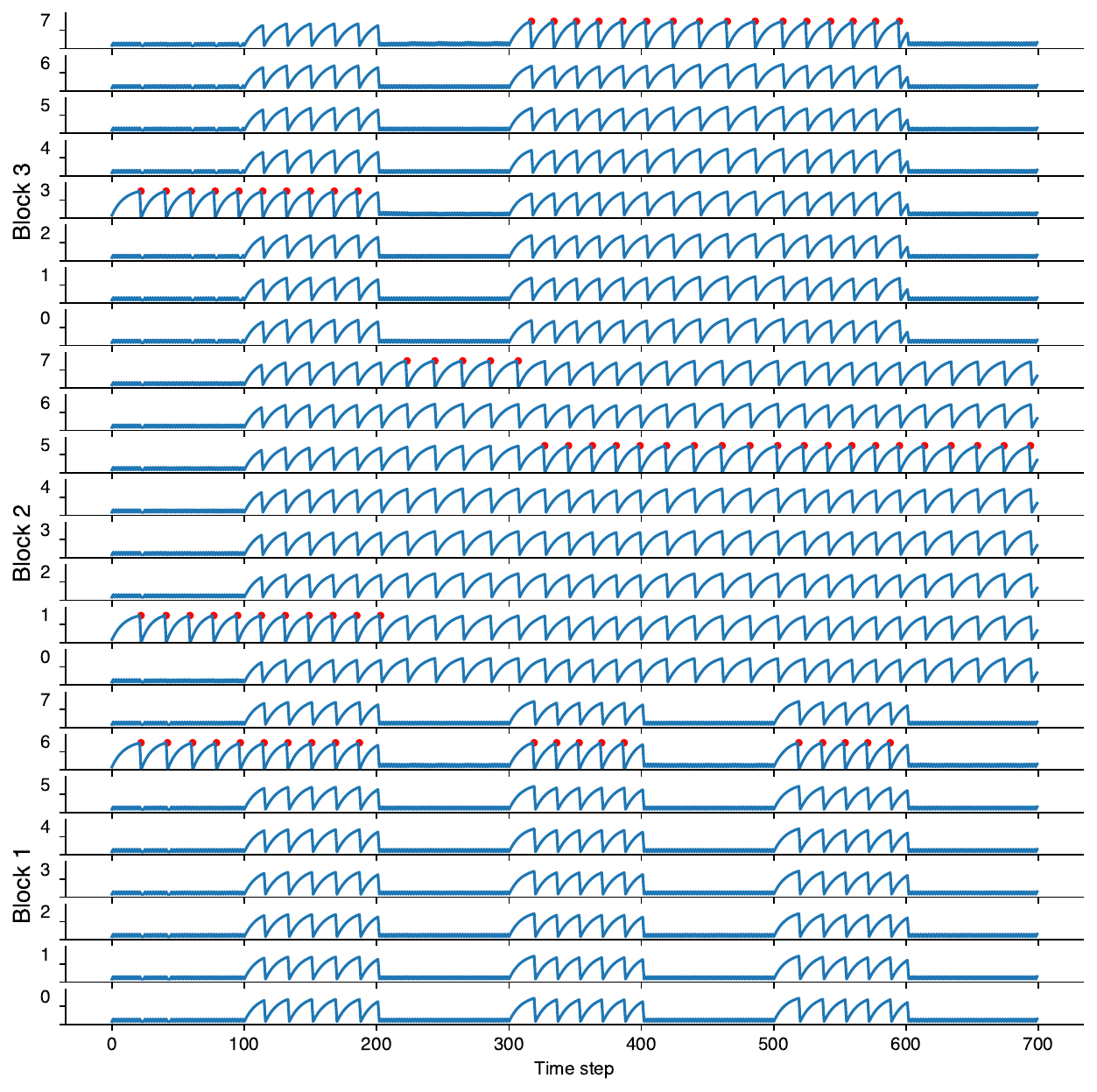}
    \caption{The winner-take-all mechanism implemented on Loihi 2. Although every neuron is receiving enough excitatory input that it will spike if unhindered, the WTA mechanism - implemented using shunting inhibition - allows only the neuron with the greatest integrated current to spike. This is an easy way to stabilize the attractor dynamics and constrains the neural activity to sparse block codes.}
    \label{fig:loihi_wta}
\end{figure}

\begin{figure}
\centering
\includegraphics[width=0.7\linewidth]{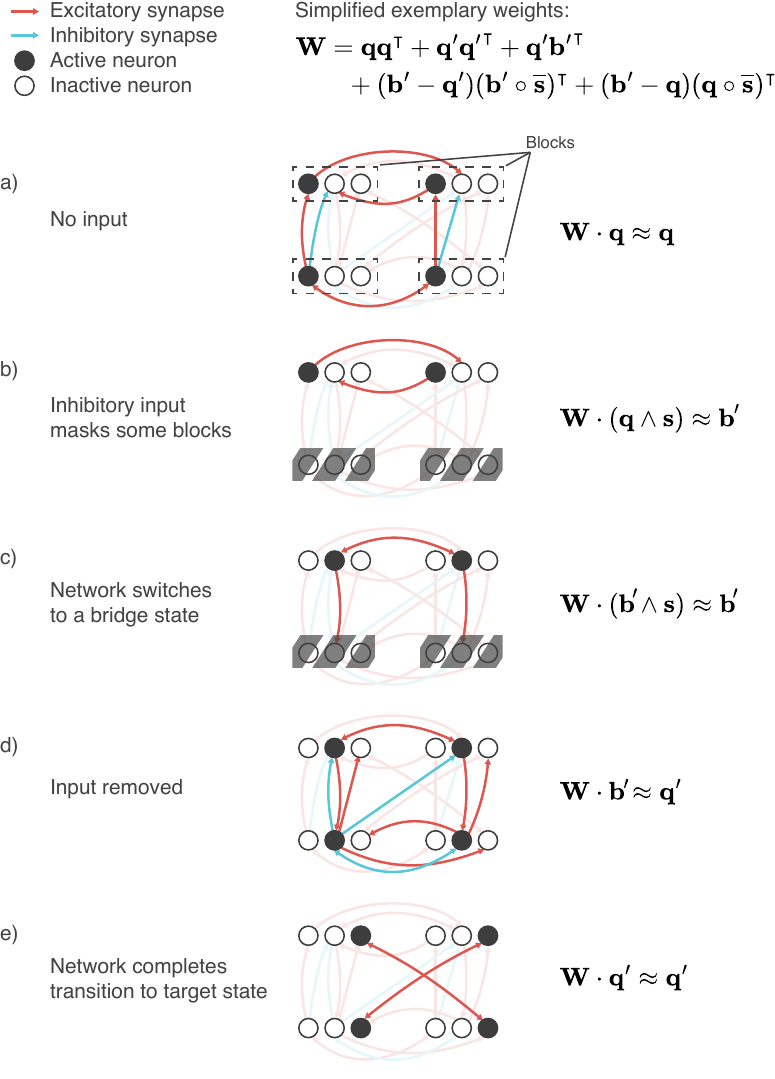}
\caption{Mechanistic description of how the constructed weight matrices result in attractor-transition dynamics. A simplified weight matrix is shown, which stores a transition from state $\vec{q}$ to $\vec{q'}$ through bridge state $\vec{b}'$.
For simplicity, the $\vec{q},\vec{b'}$ and $\vec{q'}$ hypervectors are here chosen as having the first, second and third neuron in each block ($L = 3$) be active respectively.
From one panel to the next, the neuron with the greatest input in each block becomes active, which can be calculated by summing the number of +1 excitatory and -1 inhibitory synaptic inputs.
\textbf{a)} The network starts in stable attractor state $\vec{q}$.
\textbf{b)} Applying $\vec{s}$ as a network mask ($s_i = 0/1$ for the bottom/top neuron blocks) removes some blocks from the picture, such that $\vec{q}$ ceases to be a stable state.
\textbf{c)} This causes the network to transition to the $\vec{b'}$ state, which is then stable.
\textbf{d)} The eventual removal of the input causes the network to briefly inhabit the full, unmasked $\vec{b'}$ state, which is not stable.
\textbf{e)} The network transitions to the $\vec{q}'$ state, completing the transition.
The connectivity shown here does not correspond exactly to the weight matrix above, it has been chosen for visual simplicity. In general, the states will not be orthogonal but will be randomly-generated with expected overlap $N/L^2$.
}
\label{fig:sm_transition_mechanism}
\end{figure}

\begin{figure}
\centering
\includegraphics[width=3.5in]{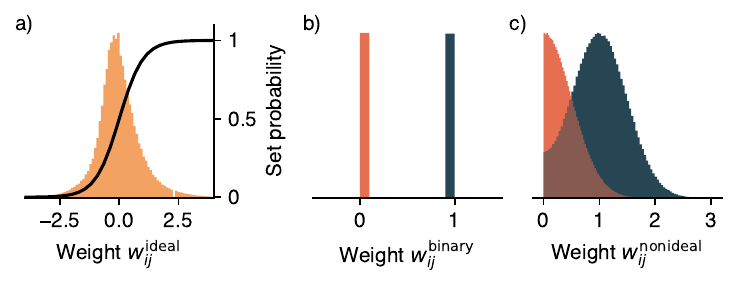}
\caption{Histograms of the recurrent weight values in simulation before and after nonidealities are applied. \textbf{a)} The ideal weights are shown in yellow, with a sigmoidal transfer function overlain. \textbf{b)} The ideal weights are stochastically binarised according to the sigmoidal set probability. \textbf{c)} The weights after adding independent Gaussian noise to each binarised weight value. The two histograms were each normalised to unit height. These nonidealities emulate that the synaptic weights are imprecise, unreliable and noisy. The weights $w_{ij}^{\text{nonideal}}$ were used in the RSNN in Fig.~\ref{fig:sim_walk}.}
    \label{fig:sim_W_distr}
\end{figure}

\begin{figure}
\centering
\includegraphics[width=0.8\linewidth]{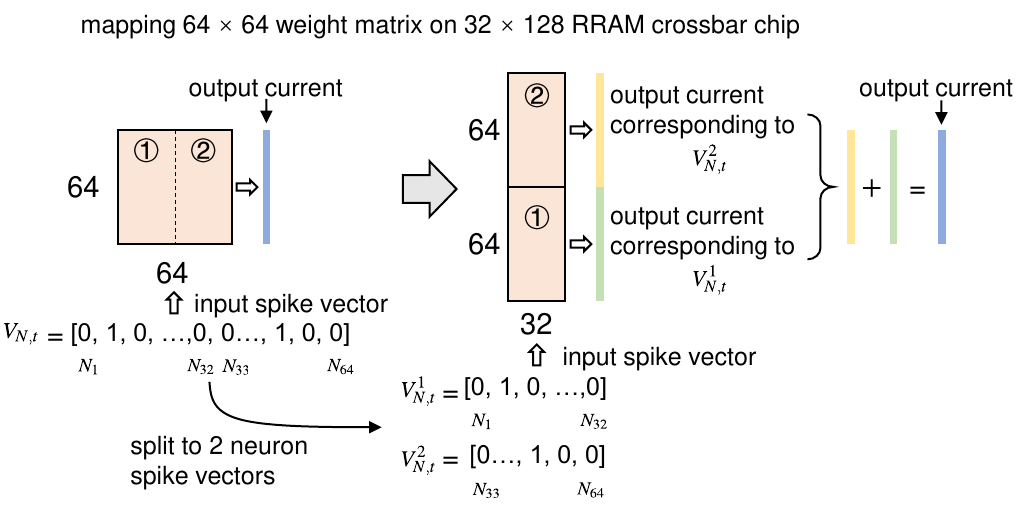}
\caption{The scheme to map the 64$\times$64 weight matrix to the 32$\times$128 RRAM crossbar chip. We split the size 64 neuron spike vector into two size 32 vectors. We then apply each vector separately as input to the RRAM crossbar one after the other and read the output currents as depicted. The sum of the two size 64 output current vectors is equivalent to applying a size 64 spike vector as input to a 64$\times$64 crossbar array (shown left). The summation of the two output current vectors is performed in software and then fed back into the RSNN simulation as postsynaptic currents.}
\label{fig:64x64_to_32x128}
\end{figure}

\begin{table}[]
\centering
\begin{tabular}{llllll}
\toprule
\textbf{Parameter} & \textbf{Description} & \textbf{First simulation}  & \textbf{Crossbar experiment} & \textbf{Big simulation} & \textbf{Loihi} \\ \midrule
$N$ & Number of neurons & 2048 & 64 & 2048 & 1024 \\
$L$ & Block length & 8 & 8 & 16 & 8\\
$\langle |w_{ij} | \rangle$ & Synaptic weight strength & $\approx \qty{0.1}{\milli\volt\per\farad}$ & $\approx \qty{4}{\milli\volt\per\farad}$ & $\approx \qty{0,2}{ \milli\volt\per\farad}$ & $\approx 45$ \\
$C$ & Membrane capacitance & \qty{1}{F} & \qty{1}{F} & \qty{1}{F} & -\\
$\tau_m$ & Membrane time constant & \qty{20}{ms} & \qty{20}{ms} & \qty{20}{ms} & 10\\
$u_\theta$ & Spiking threshold & \qty{20}{mV} & \qty{20}{mV} & \qty{20}{mV} & 1000 \\
$u_\text{rest}$ & Resting potential & \qty{25}{mV} & \qty{25}{mV} & \qty{25}{mV} & 0\\
$u_\text{reset}$ & Reset potential & \qty{0}{mV} & \qty{0}{mV} & \qty{0}{mV} & 0\\
$\tau_\text{syn}$ & Synaptic time constant & \qty{20}{ms} & \qty{20}{ms} & \qty{20}{ms} & -\\
$\tau_\text{ref}$ & Refractory period & \qty{10}{ms} & \qty{10}{ms} & \qty{10}{ms} & 0\\
$\tau_\text{readout}$ & Readout kernel timescale & \qty{10}{ms} & \qty{10}{ms} & \qty{10}{ms} & 10 \\ \bottomrule
\end{tabular}
\vspace{1em}
\caption{Parameters used in the simulation, closed-loop memristive crossbar and Loihi setups. The magnitude of the synaptic weights $\langle |w_{ij} | \rangle$ is the total amount of charge that each spike adds to a postsynaptic neuron's membrane.}
\label{tab:sim_params}
\end{table}

\FloatBarrier

\section*{Supplementary references}
\printbibliography[heading=none]

\end{refsection}

\clearpage

% \begin{refsection}
% \input{my_responses.tex}
% \end{refsection}

\end{document}